\documentclass[11pt]{article}

\PassOptionsToPackage{table}{xcolor}

\usepackage[preprint]{acl}

\usepackage{times}
\usepackage{latexsym}
\usepackage{amsmath}
\usepackage{amssymb}
\usepackage{booktabs}
\usepackage{multirow}
\definecolor{heat}{HTML}{3B6FB6}
\newcommand{\czero}{Resampled PD}
\newcommand{\std}[1]{{\scriptsize$\pm$#1}}

\usepackage[T1]{fontenc}
\usepackage[utf8]{inputenc}

\usepackage{microtype}

\usepackage{inconsolata}

\usepackage{graphicx}
\graphicspath{{figure/}}

\usepackage{xcolor}   %
\usepackage[most]{tcolorbox}
\newcommand{\hlA}[2]{{\setlength{\fboxsep}{0pt}\colorbox{apdred!#1!white}{\strut #2}}}
\newcommand{\swatch}[1]{{\setlength{\fboxsep}{0pt}%
  \colorbox{apdred!#1!white}{\rule{0pt}{6.5pt}\hspace{11pt}}}}
\newcommand{\apdlab}[1]{\makebox[3.4em][r]{\scriptsize\sffamily\textcolor{black!50}{\MakeUppercase{#1}}}\hspace{0.9em}}
\newcommand{\apdq}[1]{\noindent\apdlab{Q}\parbox[t]{\dimexpr\linewidth-4.3em}{#1}\par\vspace{2pt}}
\newcommand{\apdgen}[2]{\noindent\apdlab{#1}%
  \colorbox{black!5}{\parbox[t]{\dimexpr\linewidth-4.3em-2\fboxsep}{\strut\footnotesize #2}}%
  \par\vspace{1.5pt}}

\newtcolorbox{promptbox}[1][]{breakable, enhanced, colback=black!3, colframe=black!25,
  boxrule=0.4pt, arc=1pt, left=4pt, right=4pt, top=3pt, bottom=3pt,
  fontupper=\scriptsize\ttfamily, before upper=\raggedright,
  colbacktitle=black!12, coltitle=black, fonttitle=\footnotesize\bfseries\sffamily,
  toptitle=1.5pt, bottomtitle=1.5pt, #1}
\newcommand{\pvar}[1]{\textrm{\itshape\{#1\}}}

\title{Adaptive-Policy Distillation:\\Knowledge Injection under Information Asymmetry}

\author{Hojun Cho\textsuperscript{1} \quad Kang Eun Jeon\textsuperscript{2} \quad Jaegul Choo\textsuperscript{1} \\
  \textsuperscript{1}KAIST \quad \textsuperscript{2}DGIST \\
  \texttt{hojun.cho@kaist.ac.kr} \quad \texttt{kejeon@dgist.ac.kr} \quad \texttt{jchoo@kaist.ac.kr}}

\begin{document}
\maketitle

\begin{abstract}
On-policy distillation (OPD), which uses reverse-KL supervision on student rollouts to transfer a context-conditioned model's behavior into student model parameters for closed-book inference, has shown strong results.
However, we find that this success does not translate to knowledge injection: across two document-QA benchmarks, OPD injects document knowledge far less efficiently than simple off-policy distillation.
We trace this reversal to two failures that are structural, yet sparse across token positions.
Reverse KL weights its gradient by the student distribution and therefore barely lifts document-specific tokens the student considers unlikely; after forward KL restores this signal, student rollouts expose later positions on the target trajectory only after generating the preceding tokens correctly, so coverage deteriorates with target length.
Motivated by the concentration of teacher--student divergence at these hard positions, we propose Adaptive-Policy Distillation (APD), whose per-token gate jointly shifts the rollout toward the teacher and the objective toward forward KL where needed, withdrawing both interventions as the student learns.
Without labels or a verifier, APD improves over OPD by up to $54.6\%$, achieving the best scores of $0.558$ F1 on QASPER and $0.486$ MRR on BioASQ.
\end{abstract}

\section{Introduction}
\label{sec:intro}

Many LLM applications repeatedly answer questions using the same long documents, such as technical manuals, scientific literature, legal texts, or organization-specific knowledge.
Placing those documents in the inference context is effective~\citep{lewis2020rag}, but repeatedly processing them wastes computation and context-window capacity.
Context distillation instead injects a fixed document into model parameters: the document-conditioned model teaches the same model without the document, enabling closed-book inference without a stronger teacher, gold labels, or a verifier~\citep{askell2021general, snell2022learning, kujanpaa2025knowledge}.
Recent distillation methods increasingly replace \emph{off-policy} teacher-generated data with \emph{on-policy} supervision of the student's own sequences~\citep{song2026survey}; training on states encountered at inference reduces train--inference mismatch and often improves behavior transfer~\citep{gu2024minillm, agarwal2024onpolicy}, motivating the same extension to context distillation~\citep{ye2026opcd}.

When context distillation is used for knowledge injection, however, we observe a surprising reversal: across two document-QA datasets, pure on-policy context distillation (OPD) is consistently weaker than off-policy distillation.
Prior work separately points to weak reverse-KL supervision outside the student's effective support~\citep{minka2005divergence, wang2026notall} and inadequate rollout coverage of teacher-preferred behavior~\citep{lazaridis2026edgeopd, jia2026asymmetric}.
We ask how these two limitations interact when the teacher's only advantage is information absent from the student's input.

We study this interaction in a controlled context-distillation testbed whose documents contain random key--code mappings unavailable from the model's prior knowledge.
We vary two properties of the codes independently: token-level unpredictability and length.
For each setting, we run the full generation-source $\times$ KL-direction grid.
The results show how properties of the knowledge to be injected control the interaction between the two limitations.
Reverse KL fails regardless of generation source, making the objective the primary bottleneck; once forward KL restores the token-level signal, student rollouts become limiting as the codes grow longer.
Thus, under the information asymmetry of knowledge injection, OPD is least efficient exactly where the document contributes the most content the student cannot infer on its own.

The implication is selective rather than global: OPD should be retained where the teacher--student mismatch is modest, with intervention reserved for positions where document-specific information creates a large coverage gap.
We therefore propose \emph{Adaptive-Policy Distillation} (APD), which uses a per-token divergence gate to jointly adapt the rollout and objective.
At high-divergence positions, APD shifts sampling toward the teacher and the loss toward forward KL; otherwise, the gate remains nearly closed and training stays effectively on-policy.
As the student learns, the divergence falls and both interventions shrink toward OPD.
APD requires neither gold answers nor a verifier, yet raises OPD from below every off-policy baseline to parity with the strongest of them on our benchmarks.

Our contributions are as follows:
\begin{itemize}
\setlength{\itemsep}{0pt}
\setlength{\parsep}{0pt}
\setlength{\parskip}{0pt}
\setlength{\topsep}{2pt}

\item Using a controlled context-distillation testbed, we show how token-level unpredictability and sequence length expose distinct OPD bottlenecks along KL direction and generation source (\S\ref{sec:toy}).

\item We propose \emph{Adaptive-Policy Distillation} (APD), whose reversible per-token divergence gate jointly adapts sampling and KL direction without gold answers or a verifier (\S\ref{sec:method}).

\item Using real documents, ablations reproduce the testbed's bottleneck hierarchy and show that intervening at only a small fraction of tokens recovers most of the gain (\S\ref{sec:experiments}).
\end{itemize}

\section{Related Work}
\label{sec:related}

\subsection{Knowledge Injection via Distillation}

\citet{ye2026opcd} apply on-policy context distillation to system prompts and solution traces, whereas we study the internalization of document-specific knowledge unavailable in the student's input.
Closest to our setting, GATES distills from document-conditioned tutor traces using consensus gating~\citep{stein2026gates}, MixSD combines conditioned and unconditioned predictions at a fixed ratio~\citep{liu2026mixsd}, and EDGE-OPD guides student rollouts with a fixed teacher mixture while masking the loss using binary teacher evidence~\citep{lazaridis2026edgeopd}.
Beyond distillation, document knowledge can also be injected by synthetic continued pretraining~\citep{yang2024synthetic}.

These interventions reflect a broader difficulty of on-policy distillation: teacher-preferred content with low student probability is unlikely to be effectively learned.
Prior analyses show that student rollouts rarely reach such content~\citep{jia2026asymmetric, lazaridis2026edgeopd}, while some methods exclude teacher preferences outside the student's effective support as unteachable~\citep{wang2026notall}.
In knowledge injection, however, these low-support tokens are precisely what must be learned.
Related work further shows that forward and reverse KL can behave differently under finite training budgets despite sharing the same optimum~\citep{wu2025akl}, and that prefix source and KL direction can be varied independently~\citep{zhao2026decoupling}.
We extend this analysis to knowledge injection and use a single adaptive, reversible per-token gate to control both prefix generation and the distillation objective, rather than relying on fixed mixtures or hard masks. 

\subsection{Bridging the Teacher--Student Mismatch}

Token-level remedies for teacher--student mismatch fall into three groups.
\emph{Sampling-side} methods replace student tokens outside the teacher's top-$K$ (SKD)~\citep{xu2025skd}, let the teacher intervene when token probabilities diverge (SWITCH)~\citep{koo2025switch}, or switch to teacher generation after a moving divergence threshold (AdaSwitch)~\citep{peng2025adaswitch}.
\emph{Schedule-side} methods interpolate an intermediate teacher (TAID)~\citep{shing2025taid} or anneal a global forward-to-reverse KL mixture with a coverage-based brake (CADENCE)~\citep{kumar2026cadence}.
DistiLLM adaptively schedules student-generated data~\citep{ko2024distillm}, while DistiLLM-2 pairs different losses with teacher- versus student-generated data~\citep{ko2025distillm2}.
\emph{Loss-side} methods blend KL directions per token via probability-ratio weights on fixed data~\citep{jung2025todi} or mask unteachable tokens~\citep{wang2026notall}.
Each intervenes on sampling, scheduling, or loss alone; APD instead uses one state-dependent per-token signal for both the sampling mix and KL direction.

\begin{figure*}[t]
  \centering
  \includegraphics[width=0.9\textwidth]{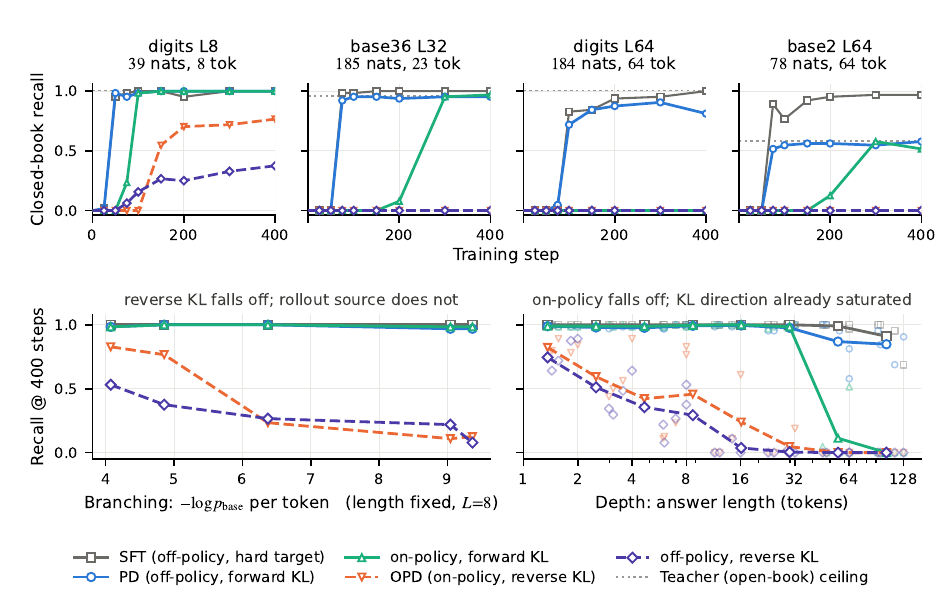}
  \vspace{-1em}
  \caption{Controlled separation of the two failure axes.
  \textbf{Top:} closed-book recall during training for four representative target cells; dotted horizontal lines show the teacher's open-book accuracy.
  \textbf{Bottom:} final recall as a function of per-token branching at fixed code length (left) and answer length across cells (right).
  Solid/dashed lines denote forward/reverse KL, faint markers are individual cells, and lines in the bottom-right panel are binned means.}
  \vspace{-1em}

  \label{fig:toy-rate}
\end{figure*}

\section{Problem Setup}
\label{sec:setup}

We consider a self-distillation problem where
the teacher is the frozen base model conditioned on the document, hence any teacher--student difference is purely informational
~\citep{stein2026gates, liu2026mixsd, lazaridis2026edgeopd}.

Let $d$ be a document to internalize and $x$ a question about it.
An autoregressive model defines $\pi(y \mid x) = \prod_{t} \pi(y_t \mid s_t)$ over responses $y=(y_1,\ldots,y_T)$, where $s_t = (x, y_{<t})$ denotes the state at step $t$.
The \emph{teacher} is the frozen base model with the document in its context, $\pi_T(\cdot \mid s_t) = \pi_0(\cdot \mid d, x, y_{<t})$.
The \emph{student} $\pi_\theta(\cdot \mid s_t)$ never sees the document and starts from the same base model without it, $\pi_0(\cdot \mid x, y_{<t})$.
The goal is for the student to reproduce the teacher's document-conditioned behavior without reading $d$.

In the distillation family studied here, the student is trained against the teacher's \emph{full} next-token distribution rather than against sampled tokens as hard labels: the loss is a per-token divergence $D$ between student and teacher, summed over the states of a sampled response~\citep{hinton2015distilling, agarwal2024onpolicy, song2026survey}.
Writing $\rho$ for the distribution of $y$, the family is
\begin{equation}
  \begin{split}
  \mathcal L(\theta)
  ={}& \mathbb E_{y \sim \rho}
  \Big[\sum_{t} D\big(\pi_\theta(\cdot\mid s_t),\,
  \pi_T(\cdot\mid s_t)\big)\Big].
  \end{split}
  \label{eq:family}
\end{equation}
Recipes within it differ along two axes.
The first is \emph{where the loss is evaluated}: the generation source $\rho$ is either a fixed corpus of teacher-generated responses, independent of the current student, on which the student is teacher-forced (\emph{off-policy}), or the student's own rollouts, resampled as the student changes (\emph{on-policy}).
The second is \emph{which divergence is minimized}: $D$ is either the forward KL $D_{\mathrm{KL}}(\pi_T \Vert \pi_\theta)$ or the reverse KL $D_{\mathrm{KL}}(\pi_\theta \Vert \pi_T)$.
Of the four resulting combinations, two are the recipes used in practice.
\textbf{PD} (per-token off-policy distillation) applies forward KL on teacher-forced states, i.e.\ word-level KD~\citep{hinton2015distilling}.
\textbf{OPD} applies token-level reverse KL to the document-conditioned teacher on the student's own rollouts~\citep{agarwal2024onpolicy}.
The remaining two, reverse KL on teacher-forced states (off-rev) and forward KL on student rollouts (on-fwd), are rarely used on their own: the two choices travel together in practice~\citep{zhao2026decoupling}.
One further reference, \textbf{SFT}, falls outside Eq.~\ref{eq:family} altogether: it fits sampled teacher tokens as hard targets without consuming the teacher logits.

\section{Why On-Policy Distillation Fails to Inject Knowledge}
\label{sec:toy}

Injection concerns knowledge the student does not have, so measuring it cleanly calls for a target the student genuinely cannot know.
We build one synthetically.
The context is a key--value store: a fixed document listing $N{=}64$ lines of the form ``\texttt{The code for \{key\} is \{code\}.}'' The codes are drawn at random, so the document is the only place they exist and no amount of prior knowledge substitutes for reading it.
For each key, we ask the probe ``\textit{What is the code for \{key\}?
Output only the code, nothing else.}''
The teacher and student share the same base model, but only the teacher receives the key--value store as document context in its system turn.
The resulting asymmetry is therefore purely informational.
To answer correctly, the student must internalize the key$\to$code mapping in its weights.
Closed-book exact-match recall over the $N$ keys therefore directly measures how many mappings have been successfully injected.

To test how the form of the knowledge affects how readily it is internalized, we vary the alphabet size ($|\Sigma| \in \{2, 10, 16, 36, 62\}$) and code length ($L \in \{2, 4, 8, 16, 32, 64, 128\}$).
These two variables control how unpredictable each code is and how many tokens must be learned in sequence.
We quantify each cell's difficulty by the mean teacher-forced, closed-book negative log-probability of its gold codes under the base model, $-\log p_{\mathrm{base}}(\text{code})$.
The resulting values range from $21$ to $632$ nats.
This controlled construction lets us vary target difficulty and length separately, which is not possible with real-data benchmarks.
On this grid we run all four cells of \S\ref{sec:setup}, with SFT as a hard-target reference and all other settings held fixed across cells.
Full experimental details are provided in Appendix~\ref{sec:appendix-synthetic}.

\paragraph{Where the signal vanishes.}
OPD minimizes the reverse KL at states the student itself visits:
\begin{equation}
  \begin{split}
  \mathcal L_{\mathrm{OPD}}
  ={}& \mathbb E_{y\sim\pi_\theta(\cdot\mid x)} \\
  & \Big[\sum_{t}
  D_{\mathrm{KL}}\big(\pi_\theta(\cdot\mid s_t)\,\Vert\,
  \pi_T(\cdot\mid s_t)\big)\Big].
  \end{split}
  \label{eq:opd}
\end{equation}
Equation~\ref{eq:opd} combines two choices that can bottleneck injection: the reverse-KL objective and the student-generated state distribution.
The full generation-source $\times$ KL-direction grid in Figure~\ref{fig:toy-rate} separates their effects.

We first examine the KL direction.
Across the figure, both reverse-KL variants lag behind their forward-KL counterparts regardless of generation source.
Even in the easiest top-left cell, both forward-KL runs saturate within $100$ steps, whereas both reverse-KL runs are still climbing at step $400$.
To see why, hold the sampled state $s$ fixed and consider the gradient with respect to the student logit of a token $a$:
\begin{equation}
  \begin{split}
  \pi_\theta(a\mid s)\Big[
    &\log\frac{\pi_\theta(a\mid s)}{\pi_T(a\mid s)} \\
    &- D_{\mathrm{KL}}\big(\pi_\theta\,\Vert\,\pi_T\big)\Big].
  \end{split}
  \label{eq:rkl-grad}
\end{equation}
The gradient is scaled by $\pi_\theta(a\mid s)$ and therefore vanishes as $\pi_\theta(a\mid s)\to0$.
Reverse KL thus provides little first-order signal for raising a low-probability token, which is its zero-forcing property~\citep{minka2005divergence, gu2024minillm, jung2025todi}.
A derivation and its assumptions are given in Appendix~\ref{sec:appendix-rkl-gradient}.
Document-specific tokens are precisely the tokens that the student initially considers unlikely and injection must raise.
The bottom-left panel makes this dependence visible.
At fixed length, a wider alphabet makes each target symbol less probable under the student; as Equation~\ref{eq:rkl-grad} predicts, the resulting reduction in gradient magnitude increasingly prevents both reverse-KL variants from learning the codes.
Both forward-KL conditions remain at ceiling, while the near-overlap of the two reverse-KL conditions shows that the generation source makes little difference in this regime.

The second bottleneck lies in the student-generated state distribution.
On-policy training places the loss on whatever the student generated.
But the student has never read the document, so at the positions that carry document-specific content it has nothing to produce the right token from; these are exactly the positions at which its rollouts leave the target.
For example, suppose the target code is \texttt{ABCD}, but the student samples \texttt{X} as its first symbol.
Every subsequent loss is then evaluated at prefixes beginning with \texttt{X}, rather than at the target prefixes \texttt{A}, \texttt{AB}, and \texttt{ABC}.
The losses after \texttt{X} cannot teach the model how to continue from the correct prefixes \texttt{A}, \texttt{AB}, and \texttt{ABC}, and therefore do not advance learning along the target code trajectory.
Training must first correct the initial error from \texttt{X} to \texttt{A} before the later target positions become exposed, so injection proceeds front to back.
Teacher forcing avoids this sequential dependency: by supplying \texttt{ABCD} during training, it places losses on the target prefixes \texttt{A}, \texttt{AB}, and \texttt{ABC} in the same update.
The bottom-right panel of Figure~\ref{fig:toy-rate} shows this coverage bottleneck most clearly: among the forward-KL conditions, gold-prefix training remains near ceiling across the full range of code lengths, whereas on-policy recall collapses beyond a few dozen tokens.
Thus, the more document-specific tokens an answer requires, the greater the disadvantage of on-policy rollouts relative to teacher forcing.

Read left to right, the top row of Figure~\ref{fig:toy-rate} shows how the two bottlenecks accumulate.
Moving from digits L8 to base36 L32 raises target difficulty and stalls both reverse-KL variants, while both forward-KL variants still reach the teacher ceiling.
At essentially the same difficulty ($185$ versus $184$ nats), lengthening the code from $23$ to $64$ tokens then collapses on-policy forward KL while PD remains strong, isolating rollout coverage.
Finally, reducing difficulty to $78$ nats at the same $64$-token length restores on-policy forward KL near the teacher ceiling, showing that difficulty shifts where the coverage wall appears but does not replace length as a distinct axis.

\begin{table*}[!t]
\centering
\small
\setlength{\tabcolsep}{3.2pt}%
\begin{tabular}{l@{\hskip 4pt}l@{\hskip 6pt}ccc@{\hskip 8pt}ccc@{\hskip 12pt}ccc@{\hskip 8pt}ccc}
\toprule
\multirow{3}{*}{Method} & \multirow{3}{*}{Loss} & \multicolumn{6}{c}{Llama-3.1-8B-Instruct} & \multicolumn{6}{c}{Qwen3-8B} \\[-1pt]
\cmidrule[0.4pt](lr){3-8}\cmidrule[0.4pt](lr){9-14}\addlinespace[-1.5pt]
& & \multicolumn{3}{c}{QASPER} & \multicolumn{3}{c}{BioASQ} & \multicolumn{3}{c}{QASPER} & \multicolumn{3}{c}{BioASQ} \\[-1pt]
\cmidrule(lr){3-5}\cmidrule(lr){6-8}\cmidrule(lr){9-11}\cmidrule(lr){12-14}\addlinespace[-1.5pt]
& & P & R & F1 & strict & len. & MRR & P & R & F1 & strict & len. & MRR \\
\midrule
\emph{no training} (closed-book) & & .158 & .164 & .148 & .121 & .199 & .151 & .140 & .250 & .157 & .115 & .209 & .151 \\
\emph{base $+$ document} (open-book) & & .756 & .860 & .764 & .733 & .769 & .748 & .699 & .805 & .701 & .678 & .762 & .713 \\
\midrule
\multicolumn{14}{l}{\emph{No teacher resampling during training}} \\
\quad SFT & CE & \cellcolor{heat!1}.475 & \cellcolor{heat!43}\underline{.622} & \cellcolor{heat!13}.485 & .412 & \cellcolor{heat!37}\underline{.488} & \cellcolor{heat!19}.446 & .352 & \cellcolor{heat!19}.506 & .373 & .310 & .463 & .373 \\
\quad PD~\citep{hinton2015distilling} & FKL & .461 & .520 & .442 & .392 & .407 & .399 & .427 & .489 & .409 & .345 & .457 & .391 \\
\quad ToDi~\citep{jung2025todi} & FKL+RKL & .474 & .539 & .460 & \cellcolor{heat!5}.418 & \cellcolor{heat!16}.457 & \cellcolor{heat!11}.435 & \cellcolor{heat!6}.430 & .492 & \cellcolor{heat!9}.415 & \cellcolor{heat!32}.364 & \cellcolor{heat!11}.470 & \cellcolor{heat!23}.407 \\
\quad GKD~\citep{agarwal2024onpolicy} & JSD & .457 & .510 & .440 & .384 & .422 & .400 & .378 & .429 & .364 & .317 & .422 & .360 \\
\quad OPD~\citep{agarwal2024onpolicy} & RKL & .411 & .449 & .390 & .317 & .333 & .324 & .309 & .378 & .302 & .244 & .323 & .276 \\
\midrule
\multicolumn{14}{l}{\emph{Teacher resampled during training}} \\
\quad \czero & FKL & \cellcolor{heat!29}.532 & \cellcolor{heat!29}.595 & \cellcolor{heat!28}.515 & \cellcolor{heat!34}\underline{.451} & \cellcolor{heat!33}.482 & \cellcolor{heat!33}\underline{.464} & \cellcolor{heat!39}\underline{.447} & \cellcolor{heat!22}.509 & \cellcolor{heat!32}.430 & \cellcolor{heat!38}\underline{.368} & \cellcolor{heat!19}.475 & \cellcolor{heat!31}\underline{.413} \\
\quad MiniLLM~\citep{gu2024minillm} & RKL & .196 & .390 & .221 & .144 & .155 & .149 & .189 & .384 & .218 & .197 & .271 & .228 \\
\quad SWITCH~\citep{koo2025switch} & skew RKL & .413 & .436 & .390 & .347 & .376 & .360 & .358 & .378 & .333 & .281 & .379 & .322 \\
\quad SKD~\citep{xu2025skd} & FKL & \cellcolor{heat!34}\underline{.543} & \cellcolor{heat!37}.610 & \cellcolor{heat!34}\underline{.527} & \cellcolor{heat!11}.425 & \cellcolor{heat!10}.447 & \cellcolor{heat!11}.435 & \cellcolor{heat!14}.434 & \cellcolor{heat!20}.507 & \cellcolor{heat!19}.422 & \cellcolor{heat!24}.360 & \cellcolor{heat!32}\underline{.484} & \cellcolor{heat!27}.410 \\
\quad AdaSwitch~\citep{peng2025adaswitch} & FKL & \cellcolor{heat!29}.533 & \cellcolor{heat!38}.613 & \cellcolor{heat!33}.524 & \cellcolor{heat!1}.413 & .432 & .421 & \cellcolor{heat!22}.438 & \cellcolor{heat!47}\underline{.529} & \cellcolor{heat!36}\underline{.433} & \cellcolor{heat!24}.359 & \cellcolor{heat!20}.476 & \cellcolor{heat!25}.408 \\
\cmidrule(lr){1-14}
\quad APD (ours) & FKL+RKL & \cellcolor{heat!50}\textbf{.574} & \cellcolor{heat!50}\textbf{.636} & \cellcolor{heat!50}\textbf{.558} & \cellcolor{heat!50}\textbf{.469} & \cellcolor{heat!50}\textbf{.508} & \cellcolor{heat!50}\textbf{.486} & \cellcolor{heat!50}\textbf{.453} & \cellcolor{heat!50}\textbf{.531} & \cellcolor{heat!50}\textbf{.442} & \cellcolor{heat!50}\textbf{.376} & \cellcolor{heat!50}\textbf{.495} & \cellcolor{heat!50}\textbf{.426} \\
\bottomrule
\end{tabular}
\caption{Closed-book knowledge injection, mean over 20 single-document units at the final checkpoint (standard deviations in Appendix).
\textbf{Best} and \underline{second best} among trained methods in each column; cells above the column median are shaded in proportion to their score.
Loss: FKL/RKL = forward/reverse KL, skew RKL as in \citet{ko2024distillm}, JSD = Jensen--Shannon, CE = cross-entropy on teacher samples; FKL+RKL combines both directions per token.}
\vspace{-1em}
\label{tab:main}
\end{table*}

\section{Adaptive-Policy Distillation}
\label{sec:method}

The analysis above suggests a selective remedy.
At document-specific positions where the student assigns little probability to teacher-preferred tokens, the objective should shift toward forward KL to restore the missing gradient, while the rollout should shift toward the teacher to keep later target positions reachable.
Applying these interventions everywhere, however, would discard OPD's main advantage: training on the student-generated states encountered at inference.
We therefore propose \emph{Adaptive-Policy Distillation} (APD), which intervenes only where teacher and student disagree and otherwise reduces to OPD.
APD implements this adaptation with a single gating signal: the per-token divergence between teacher and student at the current state,
\begin{equation}
  \kappa_t
  = D_{\mathrm{KL}}\big(\pi_T(\cdot\mid s_t)\,\Vert\,
    \pi_\theta(\cdot\mid s_t)\big),
  \label{eq:gate}
\end{equation}
from which the gate
\begin{equation}
  g_t = 1 - e^{-\kappa_t/\tau} \in [0,1)
  \label{eq:gatedef}
\end{equation}
is formed, where $\tau$ is the method's single hyperparameter and $\kappa_t$ is treated as a constant during backpropagation.
The temperature $\tau$ controls the gate's selectivity: for a fixed divergence, increasing $\tau$ keeps $g_t$ closer to zero, so moderate disagreements continue to use OPD and only large mismatches receive substantial teacher intervention.
Which mismatches open the gate is determined by the direction of $\kappa_t$.
Because teacher and student share weights and differ only in whether they receive the document, probability mass that is high under the teacher but low under the student identifies tokens likely to carry document-specific information missing from the student.
The forward divergence $D_{\mathrm{KL}}(\pi_T\Vert\pi_\theta)$ weights precisely this teacher-preferred mass.
Accordingly, $g_t$ is \emph{open} ($g_t \to 1$) at large teacher--student coverage gaps and remains near \emph{closed} ($g_t \approx 0$) when the divergence is small relative to $\tau$, allowing moderate differences to be handled by OPD.

The gate intervenes twice, with the same signal.
During rollout, the next token is drawn from the per-token mixture
\begin{equation}
  y_t \sim m_t
  = (1-g_t)\, \pi_\theta(\cdot\mid s_t)
  + g_t\, \pi_T(\cdot\mid s_t),
  \label{eq:sampling}
\end{equation}
so generation remains effectively on-policy wherever $\kappa_t$ is small relative to $\tau$, and shifts substantially toward the teacher only at high-divergence positions; no token is ever discretely handed over.
In the loss, the same gate blends the two KL directions at every state of the resulting rollout:
\begin{equation}
  \begin{split}
  \mathcal L_{\mathrm{APD}}
  = \mathbb E_{y \sim m}
  \Big[\sum_t
    & (1-g_t)\, D_{\mathrm{KL}}\big(\pi_\theta \Vert \pi_T\big) \\[-2pt]
    & + g_t\, D_{\mathrm{KL}}\big(\pi_T \Vert \pi_\theta\big)
  \Big],
  \end{split}
  \label{eq:loss}
\end{equation}
with all divergences evaluated at $s_t$.

APD uses only the teacher logits already required by OPD, with no labels, verifier, or additional teacher evaluations.
The gate and mixture add only elementwise operations, so the FLOP count is unchanged and matched timing runs show wall-clock parity with OPD when memory is sufficient (Appendix~\ref{sec:appendix-rollouts}).
The added cost is memory: the teacher generates alongside the rollout, so its KV cache remains resident during training rollout generation.

\section{Experiments}
\label{sec:experiments}

\label{sec:exp-data}

We evaluate document knowledge injection on two generative QA benchmarks requiring document-specific content: QASPER~\citep{dasigi2021qasper}, information-seeking QA over full NLP papers (up to $\sim$9k tokens; 16{,}384-token context), scored on answerable questions with token-level Answer-F1 and its precision and recall components; and BioASQ~\citep{krithara2023bioasq}, biomedical factoid QA over full PMC articles, scored by strict and lenient accuracy and mean reciprocal rank (MRR) over up to five ranked answers.
Dataset details are given in Appendix~\ref{sec:appendix-datasets}.
For each benchmark, we use twenty documents as separate injection units: every method is trained independently on QA about each document and evaluated closed-book.
All methods, including APD, are trained on two base models, Llama-3.1-8B-Instruct~\citep{grattafiori2024llama3} and Qwen3-8B~\citep{yang2025qwen3} in non-thinking mode, referred to below as Llama-3.1 and Qwen3.
We train every method for 10 epochs with global batch size 64, AdamW ($10^{-5}$), and seed 42, and evaluate the final checkpoint.
All other training, rollout, and decoding settings are held fixed across all methods.
For APD, $\tau$ is selected once per model on ten held-out BioASQ development documents and used for both datasets.
Training details and the selection procedure are given in Appendix~\ref{sec:appendix-details}.
Tables report per-method means over documents (absolute scores).
Since every method trains on the same document set, significance is assessed with paired $t$-tests across documents (Appendix~\ref{sec:appendix-significance}).

\subsection{Training and Evaluation Data}
\label{sec:exp-datagen}
Because each benchmark provides only one to six questions per document, we construct document-specific training and evaluation sets from the source documents.
For training, Qwen3-235B generates questions from each document, and the document-conditioned base model generates the corresponding answers, yielding $0.6$--$1.3$k QA pairs per document.
For evaluation, we generate a held-out set of approximately 100 questions and reference answers per document using GPT-5.4~\citep{openai2026gpt54}, with the benchmark's human-written examples provided as demonstrations to preserve its question and answer style.
To validate the evaluation set, we compare closed-book and open-book performance of the untrained model.
Providing the source document increases QASPER F1 from $.148$ to $.764$ and BioASQ MRR from $.151$ to $.748$ for Llama-3.1, and from $.157$ to $.701$ and $.151$ to $.713$ for Qwen3, indicating that the generated questions are answerable from the document but largely inaccessible without it.
The generation procedure and all prompts are given in Appendices~\ref{sec:appendix-datagen} and~\ref{sec:appendix-prompts}.

\subsection{Baselines}
\label{sec:exp-baselines}

SFT and PD generate a corpus from the teacher once before training and reuse it in every epoch.
APD instead draws tokens from the teacher wherever its gate opens, so its teacher samples change at every step.
A fixed corpus thus repeats the same teacher answers in every epoch, whereas resampling adds new teacher information each epoch.
Prior token-level methods similar to APD compare against fixed-corpus or student-rollout baselines, but not against a resampled teacher corpus~\citep{xu2025skd, koo2025switch, peng2025adaswitch}.
To separate the effect of resampling, we group the baselines by this property, as in Table~\ref{tab:main}.

\paragraph{No teacher resampling.} Besides SFT and PD, this group includes methods that generate no teacher samples at all, such as OPD.
It also includes GKD~\citep{agarwal2024onpolicy}, which mixes the fixed corpus and student rollouts at a fixed rate ($\lambda{=}0.5$) under the Jensen--Shannon divergence, and ToDi~\citep{jung2025todi}, which blends forward and reverse KL per vocabulary entry on the fixed corpus.

\paragraph{Teacher resampled during training.} We refer to PD with teacher responses resampled at every step instead of a fixed corpus as \emph{\czero}.
It is equivalent to APD with the gate fully open ($g_t \equiv 1$).
Since injection depends on the teacher exposing content the student cannot produce (\S\ref{sec:toy}), resampling's advantage over PD grows with the number of training epochs (Appendix~\ref{sec:appendix-epochs}).
We also evaluate resampled counterparts of SFT and GKD (Appendix~\ref{sec:appendix-resample}).
Four published methods also resample the teacher during training: SKD~\citep{xu2025skd}, which replaces student tokens outside the teacher's top-$K$ and applies forward KL; SWITCH~\citep{koo2025switch}, which lets the teacher intervene when token probabilities diverge; AdaSwitch~\citep{peng2025adaswitch}, which switches to the teacher once the divergence exceeds a moving threshold; and MiniLLM~\citep{gu2024minillm}, which mixes the teacher into every token at a fixed rate and optimizes sequence-level reverse KL by policy gradient.
Baseline-specific hyperparameters follow the original papers.
Deviations required by our setting are given in Appendix~\ref{sec:appendix-gate-hparams}.

\begin{table}[t]
\centering
\small
\setlength{\tabcolsep}{4pt}%
\begin{tabular}{l*{3}{>{\centering\arraybackslash}p{3.6em}}}
\toprule
& \multicolumn{3}{c}{loss} \\
\cmidrule(lr){2-4}
sampling & RKL & gated & FKL \\
\midrule
student & .324 & \cellcolor{heat!18}.472 & .450 \\
gated & .309 & \cellcolor{heat!50}\textbf{.486} & \cellcolor{heat!34}.479 \\
teacher & .310 & \cellcolor{heat!39}.481 & .464 \\
\bottomrule
\end{tabular}
\caption{Decomposing the two axes of APD (Llama-3.1 on BioASQ, MRR, mean over 20 documents).
Rows set the source of the rollout, which is the student, the gated mixture of APD ($\tau{=}0.3$), or the teacher.
Columns set the loss, which is reverse KL (RKL), the gated combination of APD, or forward KL (FKL).
The diagonal is OPD, APD, and \czero.
Cells above the median are shaded in proportion to their score.}
\vspace{-1em}
\label{tab:axes}
\end{table}

\begin{table}[t]
\centering
\small
\setlength{\tabcolsep}{2.6pt}%
\begin{tabular}{lcccccc}
\toprule
\multirow{2}{*}[-2.3pt]{Setting} & \multicolumn{3}{c}{gate $\bar g$} & \multicolumn{3}{c}{BioASQ} \\
\cmidrule(lr){2-4}\cmidrule(lr){5-7}
& start & mean & end & strict & lenient & MRR \\
\midrule
$\tau{\to}\infty$ {\scriptsize(OPD)} & .00 & .00 & .00 & .317 & .333 & .324 \\
$\tau{=}10$ & .03 & .02 & .02 & .424 & .472 & .444 \\
$\tau{=}3$ & .09 & .06 & .05 & .453 & .489 & .469 \\
$\tau{=}1$ & .22 & .14 & .09 & .465 & .510 & .485 \\
$\boldsymbol{\tau{=}0.3}$ {\scriptsize\textbf{(APD)}} & \textbf{.44} & \textbf{.26} & \textbf{.19} & \textbf{.469} & \textbf{.508} & \textbf{.486} \\
$\tau{=}0.1$ & .56 & .38 & .30 & .463 & .493 & .476 \\
$\tau{=}0.01$ & .74 & .60 & .54 & .457 & .486 & .469 \\
$\tau{\to}0$ {\scriptsize(\czero)} & 1.00 & 1.00 & 1.00 & .451 & .482 & .464 \\
\midrule
uniform 0.25 & .25 & .25 & .25 & .426 & .468 & .445 \\
\bottomrule
\end{tabular}
\caption{Gate temperature and a constant gate (mean over 20 documents, final checkpoint).
$\bar g$ is the mean gate value, i.e.\ the fraction of teacher mass injected, for the untrained student, averaged over training, and at the end of training.
\emph{uniform 0.25} replaces the per-token gate by a constant matched to the mean gate of APD, applied to both axes.
Our setting is in \textbf{bold}.}
\vspace{-1em}
\label{tab:ablation}
\end{table}

\subsection{Main Results}
\label{sec:exp-main}

Table~\ref{tab:main} compares APD with the baselines on both models and datasets.
We discuss the primary metrics, F1 for QASPER and MRR for BioASQ.
Appendix~\ref{sec:appendix-full} adds standard deviations and significance tests for every metric.
We call a difference significant when a two-sided paired $t$-test over the twenty documents gives $p<.10$, that is, at the $90\%$ confidence level.

\paragraph{On-policy distillation falls below every fixed-corpus method.}
OPD samples only from the student and trains with reverse KL.
In all four settings it scores below every method that trains on a fixed teacher corpus, by $.050$ to $.084$ against the lowest of them.
Under information asymmetry, distilling on the student's own rollouts thus injects less than imitating answers that the teacher generated once before training, as our analysis in \S\ref{sec:toy} predicts.

\paragraph{Resampling the teacher helps with forward KL.}
The methods that resample the teacher with a forward-KL term in the loss, \czero{}, SKD, AdaSwitch, and APD, are on average above the fixed-corpus baselines in every setting, by $.031$ to $.074$.
MiniLLM and SWITCH also resample the teacher but do not gain from it.
Both train with reverse-KL objectives and fall below every fixed-corpus method.
This is the pattern of the testbed in \S\ref{sec:toy}, where the source of the training sequences matters only once the loss is forward KL.
Under information asymmetry, seeing teacher sequences repeatedly is thus not enough.
The loss must raise the student's low token probabilities.

\paragraph{APD is the best method in every setting.}
APD scores highest in all four settings, above the resampled forward-KL baselines as well as the fixed-corpus ones.
Adding its gate raises OPD by $.140$ to $.168$ in every setting ($t \geq 6.5$).
SKD and AdaSwitch, whose rules decide token by token whether the student or the teacher generates, are on par with \czero{} in three settings and significantly below it on Llama-3.1 BioASQ ($-.029$ and $-.043$ MRR), and Appendix~\ref{sec:appendix-gate-hparams} examines how their sampling rules behave in our setting.
APD, in contrast, is significantly above \czero{} in three of the four settings.
More broadly, it is significantly above every baseline in every setting except three comparisons on BioASQ, against \czero{} for Llama-3.1 and against SKD and AdaSwitch for Qwen3, which it still exceeds by $.016$ to $.022$ MRR.

\subsection{Ablations and Analysis}
\label{sec:exp-ablation}
All ablations use Llama-3.1 on BioASQ with the twenty test documents of Table~\ref{tab:main}.

\begin{figure}[!t]
\centering
\includegraphics[width=0.95\columnwidth]{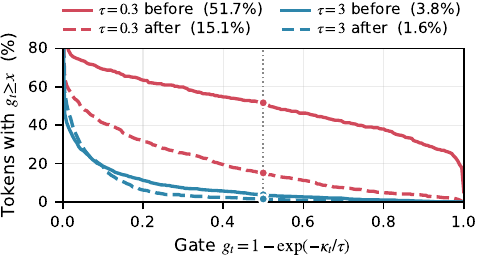}
\caption{Per-token gate distributions over mixture rollouts of Llama-3.1 on one BioASQ document, before and after 10 epochs of APD at $\tau{=}0.3$ and $\tau{=}3$.
Each curve gives the share of tokens with $g_t\geq x$.
Parentheses give the share with $g_t>0.5$.}
\vspace{-1em}
\label{fig:gate-dist}
\end{figure}

\paragraph{The loss comes first, then sampling.}
Table~\ref{tab:axes} sets the sampling and loss axes independently.
With reverse KL, every sampling choice stays at the level of OPD ($.309$--$.324$).
Letting the teacher generate does not help as long as the loss cannot raise the student's low token probabilities.
For every sampling choice, forward KL raises MRR over reverse KL by $.126$ to $.171$, and the gated loss raises it further by $.007$ to $.021$.
Once the loss is forward KL or gated, sampling also orders the cells, with student sampling lowest, teacher sampling next, and gated sampling highest.
Gating is thus the best choice on each axis once reverse KL is excluded, and full APD, which gates both, scores highest ($.486$).
This matches the pattern of the testbed (\S\ref{sec:toy}) and of the main results.

\paragraph{The gate temperature.}
Table~\ref{tab:ablation} varies $\tau$ between OPD, where the gate is always closed ($\tau{\to}\infty$), and \czero{}, where it is always open ($\tau{\to}0$).
MRR rises steeply as soon as the gate begins to open.
At $\tau=10$, where the teacher contributes only $2$--$3\%$ of the sampling mass, MRR rises from $.324$ to $.444$, and at $\tau=3$, with about $6\%$, to $.469$.
Beyond this point MRR changes little, peaking at the selected $\tau=0.3$ ($.486$), staying within $.010$ of it for $\tau$ from $0.1$ to $1$, and ending at $.464$ when the gate is always open.
The same pattern holds for both models and both datasets (Appendix Table~\ref{tab:gate-mass}).
In every setting, $\tau=10$ already recovers $52$--$74\%$ of the gain of the selected $\tau$ over OPD, and $\tau=3$ recovers $77$--$98\%$.
Intervening at a small fraction of tokens thus removes most of the deficit of OPD, and additional teacher mass refines the rest.

\paragraph{Where versus how much.}
To separate where the gate opens from how much teacher mass it injects, the last row of Table~\ref{tab:ablation} replaces the per-token gate on both axes with a constant gate of $0.25$, matched to the average gate of APD ($.26$).
This constant gate reaches $.445$, significantly below APD ($-.042$, $t=-3.67$) and also below \czero{}.
It does no better than $\tau=10$ ($.444$), which injects about a tenth of that mass at the positions of largest divergence.
Figure~\ref{fig:gate-dist} shows how widely the gate opens.
At $\tau=3$, the gate exceeds $0.5$ at only $3.8\%$ of tokens before training and $1.6\%$ after, yet APD still reaches $.469$ MRR, close to the $.486$ of $\tau=0.3$ and far above OPD.
At $\tau=0.3$, the gate opens at $51.7\%$ of tokens before training and narrows to $15.1\%$ during training.
Most of the gain of APD therefore comes from a small set of high-divergence positions, and the wider opening at $\tau=0.3$ adds the rest.
Appendix~\ref{sec:appendix-gate-example} visualizes these positions on two rollouts.
The gate opens on document-specific content that the student does not yet cover and closes once the student covers it.

\section{Conclusion}
\label{sec:conclusion}
We analyzed why on-policy distillation becomes inefficient for knowledge injection, identifying weak reverse-KL supervision and limited student-rollout coverage.
Based on this analysis, we proposed APD, which selectively adapts both the loss and rollout where teacher--student divergence is large.
Our experiments show that the two failures are addressed in order, since sampling from the teacher helps only once the loss can raise the low token probabilities of the student, and that intervening at a small set of high-divergence positions already removes most of the deficit of OPD.

\section*{Limitations}

We study knowledge injection, where the teacher holds information that the student cannot infer, so their distributions differ sharply on the content to be injected.
We therefore do not test general distillation, where the student already produces a distribution close to that of the teacher, and other bottlenecks may dominate there.
Our experiments cover two 8B backbones and two document-QA benchmarks, so how the two bottlenecks change with model scale and domain remains open.
Most real-data evaluation questions and references are model-generated.
Open-book checks and the smaller human-written question sets support their validity, but generator artifacts may still affect absolute scores or method rankings.
Our comparisons measure learning efficiency at fixed budgets---400 steps in the testbed and 10 epochs on real documents---rather than asymptotic injection capacity, and additional compute or method-specific tuning could alter the ordering.
The controlled key--code testbed deliberately removes linguistic structure and uses one training configuration, so it isolates the two mechanisms but does not establish how frequently they dominate natural knowledge injection.
Finally, APD has similar FLOPs and wall-clock time to OPD when memory is sufficient, but because the teacher must stay resident during generation, it cannot offload one model while the other runs, which may limit scaling to larger models or longer contexts.

\bibliography{custom}

\appendix

\section{Reverse-KL Gradient at Low Student Probability}
\label{sec:appendix-rkl-gradient}

We derive the logit gradient used in \S\ref{sec:toy}.
At a fixed state $s$, write $p_a=\pi_\theta(a\mid s)$ and
$q_a=\pi_T(a\mid s)$.
The reverse KL is
\begin{equation}
  R(p,q)=\sum_b p_b\log\frac{p_b}{q_b}.
  \label{eq:appendix-rkl}
\end{equation}
For student logits $z_a$, the softmax Jacobian is
$\partial p_b/\partial z_a=p_b(\mathbf 1[a=b]-p_a)$.
Treating the sampled rollout state as fixed during backpropagation gives
\begin{align}
  \frac{\partial R}{\partial z_a}
  &= \sum_b p_b(\mathbf 1[a=b]-p_a)
     \left(\log\frac{p_b}{q_b}+1\right) \notag\\
  &= p_a\left[\log\frac{p_a}{q_a}-R(p,q)\right],
  \label{eq:appendix-rkl-gradient}
\end{align}
which is Eq.~\ref{eq:rkl-grad}.

For a finite vocabulary, the teacher softmax has full support, so $q_a>0$
and $R(p,q)$ remains bounded as $p_a\to0^+$.
Using $\lim_{p\to0^+}p\log p=0$, both terms in
Eq.~\ref{eq:appendix-rkl-gradient} vanish as $p_a\to0^+$, and hence
\begin{equation}
  \lim_{p_a\to0^+}
  \left|\frac{\partial R}{\partial z_a}\right|=0.
  \label{eq:appendix-rkl-limit}
\end{equation}
Thus the direct first-order signal for increasing a token vanishes with its
student probability.
This is a local, finite-budget statement about optimization rate, not a claim
that reverse KL has a different optimum or that the token is asymptotically
unlearnable through other states.

\section{Datasets}
\label{sec:appendix-datasets}

All documents, questions, and answers used in our experiments are in English.

\subsection{Benchmarks and Metrics}

We use QASPER and BioASQ because they provide long, knowledge-intensive documents in specialized fields (science and medicine), human-written question--answer pairs that can serve as demonstrations for generating evaluation questions, and established metrics.

\paragraph{QASPER.} QASPER contains information-seeking questions about NLP papers.
Each document is a full paper of up to approximately $9$k tokens and fits within our $16{,}384$-token context window.
Answer-F1 is token overlap between the generated and gold answer after standard normalization, maximized over gold references.
We also report precision and recall because closed-book answers vary in length: recall tracks whether the injected content surfaces, while precision tracks whether it is stated without padding.
We score only answerable questions because success on an unanswerable question measures abstention rather than retention of document content.

\paragraph{BioASQ.} BioASQ contains biomedical factoid questions, each paired
with the full text of the PMC article that answers it.
The model returns up to five ranked candidates: strict accuracy requires the top candidate to exactly match the gold answer or a listed synonym, lenient accuracy accepts a match anywhere in the list, and MRR uses the rank of the first correct candidate.

\subsection{Data Generation}
\label{sec:appendix-datagen}

Each dataset contributes twenty test single-document units, QASPER papers and PMC articles paired with BioASQ factoid questions, and BioASQ also contributes ten development units for selecting $\tau$ (Appendix~\ref{sec:appendix-tau}).
A unit corresponds to one training run and contributes one observation to the paired tests reported throughout.
Every unit carries three question sets: the benchmark's human-written questions (two to six per QASPER paper, $72$ in total over the test units of which $62$ are answerable; one per BioASQ article, $20$ in total), a training set, and a generated evaluation set.
All generation prompts are reproduced in Appendix~\ref{sec:appendix-prompts}.

\begin{table*}[t]
\centering
\small
\setlength{\tabcolsep}{4pt}%
\renewcommand{\arraystretch}{1.15}%
\begin{tabular}{@{}p{0.145\textwidth}p{0.39\textwidth}p{0.42\textwidth}@{}}
\toprule
& Question & Answer \\
\midrule
\multicolumn{3}{@{}l}{\emph{QASPER, one unit (paper 1806.09103)}} \\
training & Why does the performance of deep neural networks drop significantly when the ground-truth answer contains out-of-vocabulary words in cloze-style reading comprehension? & The performance of deep neural networks drops significantly when the ground-truth answer contains out-of-vocabulary (OOV) words in cloze-style [\ldots] \emph{[397 words]} \\
training & Why might the element-wise product (mul) be superior in combining embeddings? Answer in 15 words or fewer. & Element-wise product captures interactions and reduces distribution differences between embeddings. \\
generated eval & what mechanism is introduced by Cui2016Attention? & attended attention (AoA) \\
human-written & how are rare words defined? & low-frequency words \\
\midrule
\multicolumn{3}{@{}l}{\emph{BioASQ, one unit (article PMC3472786)}} \\
training & True or false: Mutations in SWI/SNF subunits have been associated with cancer in humans. & True. Mutations in SWI/SNF subunits have indeed been associated with cancer in humans. The text explicitly states: ``Mutations in SWI/SNF subunits (SNF5, BRG1, BRM) have been detected in [\ldots] \emph{[56 words]} \\
training & Despite chromatin restricting access to DNA, what two general roles do ATP-dependent chromatin remodeling complexes play in nucleotide excision repair? Provide your answer in two short phrases. & Facilitate DNA accessibility by remodeling nucleosomes; Enhance recruitment and assembly of NER factors at damage sites \\
generated eval & Which gene locus required Gcn5 for UV-induced H3 acetylation at K9 and K14 and efficient NER? & MFA2 \\
human-written & Chromatin is tightly packed into what structure inside the nucleus? & Nucleosome \\
\bottomrule
\end{tabular}
\caption{Examples from the three question sets of one unit per dataset.
The examples illustrate how an explicit answer-format cue can elicit a short training answer, whereas an open-ended question can elicit a much longer one.
Generated evaluation questions are prompted to match the form and answer length of the human-written examples.
Long training answers are truncated at the ellipsis, with their full length in brackets.}
\label{tab:examples}
\end{table*}

\paragraph{Training questions.} Questions are generated by Qwen3-235B-A22B-Instruct-2507 at temperature $0.9$.
The document is first split along its own markdown structure into passages, and the generator is then queried at three scopes: \emph{single}, one passage at a time (target $2{,}400$ questions per document); \emph{cross}, a sliding group of passages with the requirement that an answer draw on two or more of them ($1{,}200$); and \emph{full}, the whole document, for questions about its overall aim and central findings ($450$).
Each request asks for a number of questions proportional to the passage length (between $2$ and $15$) and returns a JSON array of question/evidence pairs, the evidence being a verbatim quote supporting the answer.
The system prompt is domain-neutral: it says only ``article'', names no field, and is identical for both datasets.
It requires the questions to be self-contained, forbidding meta-reference to the prompt, and to vary widely in form.
The collected questions are then deduplicated per document by MinHash LSH over character 3-grams at Jaccard threshold $0.45$, and any question that still matches a list of meta-reference phrases, such as ``the passage'' or ``the author'', is dropped.
Because the hashing procedure is approximate, some near-duplicates may remain in the training set.

\paragraph{Training answers.} Answers are generated by the base model itself,
greedily (temperature $0$, up to $512$ tokens), from the document in a \texttt{<context>} block followed by the question and with no system prompt: this is exactly the input the teacher $\pi_T$ receives during training (\S\ref{sec:setup}).
Thus, the off-policy targets and the on-policy teacher use the same checkpoint and document-conditioned input.
Empty answers are dropped, along with the occasional runaway generation, leaving between $616$ and $1{,}286$ QA pairs per test unit, of which about two thirds are single-scope, most of the rest cross-passage, and a small remainder full-document.

\paragraph{Evaluation set.} The evaluation set is generated per document by
GPT-5.4 through its API, in a single call supplying the article, up to five of that document's human-written (question, answer) pairs as style examples, and a request for $100$ diverse, non-redundant pairs, each naming a specific entity and answerable only from the article.
For BioASQ the prompt additionally constrains the answer to a single named entity or short canonical term: the factoid metric is exact match against a reference string, so a descriptive gold phrase would make the score reflect phrasing rather than knowledge.
Returned pairs are retained only if the answer occurs in the article and the question has a word-level Jaccard overlap below $0.45$ with every question already accepted for the same document.
Unlike the approximate hashing used for the training set, this exact comparison leaves no pair above the threshold.
Between $78$ and $116$ questions per test unit survive.
Table~\ref{tab:examples} shows examples drawn from all three sets of one unit per dataset.

\section{Experimental Details}
\label{sec:appendix-details}

\subsection{Synthetic Testbed}
\label{sec:appendix-synthetic}

The key--code testbed of \S\ref{sec:toy} uses Qwen3-1.7B as both teacher and student, with the teacher frozen and reading the key--value document.
Every cell shares one configuration: global batch $64$, $400$ steps, learning rate $5{\times}10^{-6}$, seed $42$.
Closed-book recall is measured by exact string match over the $N{=}64$ keys.

\subsection{Real-Document Experiments}
\label{sec:appendix-realdata}

\paragraph{Training.} Student and teacher are the same base checkpoint, Llama-3.1-8B-Instruct or Qwen3-8B.
The teacher is frozen and reads the document, whereas the student never does.
For Qwen3, thinking mode is disabled throughout via the chat template.
Parameters are held in BF16 with FP32 optimizer states under DeepSpeed ZeRO-2 across two B200 GPUs.
We use AdamW (learning rate $10^{-5}$, $\beta = (0.9, 0.999)$, $\epsilon = 10^{-8}$, weight decay $0.01$), gradient clipping at $1.0$, a cosine schedule with warmup ratio $0.03$, and a maximum sequence length of $16{,}384$ tokens, which covers the longest document in either corpus.
The global batch size of $64$ is reached by gradient accumulation.
Every method is trained for $10$ epochs and evaluated at the final checkpoint.
All KL and Jensen--Shannon divergences are computed over the full vocabulary.
No top-$k$ truncation is applied to these objectives.

\paragraph{Rollouts.}\label{sec:appendix-rollouts} On-policy methods generate with vLLM colocated with the
trainer on the same two GPUs, and the engine's weights are refreshed from the current student at every step, so the rollouts are on-policy in the strict sense.
Completions are capped at $512$ tokens.
Rollouts are sampled at temperature $1.0$ with no nucleus or top-$k$ truncation, except for SKD and AdaSwitch, which use the sampling settings of their original papers (Appendix~\ref{sec:appendix-gate-hparams}).
APD additionally maintains a second engine for the teacher.
The teacher reads the document and the student does not.
At every position, both engines compute next-token distributions on the same prefix, the gate of Eq.~\ref{eq:gate} mixes them, and the sampled token is appended to both.
Both engines run with chunked prefill and prefix caching enabled.
These change how the forward pass is scheduled, not what it returns.
Table~\ref{tab:timing} compares the training cost of OPD and APD.
The rollouts of APD take longer because the teacher decodes alongside the student, but APD reuses these teacher logits in the loss instead of running a separate teacher forward pass.
In addition, the rollouts of APD are $15$--$33\%$ shorter than those of OPD, as they are partly sampled from the teacher, which reads the document.
Its step time is therefore comparable to or lower than that of OPD.
The cost is that both engines must stay resident while APD generates, whereas OPD can generate with the student and score with the teacher in separate phases and offload whichever model is idle.

\begin{table}[t]
\centering
\small
\setlength{\tabcolsep}{2.7pt}%
\begin{tabular}{lcccccc}
\toprule
& & \multicolumn{3}{c}{Time per step (s)} & & \\
\cmidrule(lr){3-5}
& & Rollout & Update & Total & Length & Memory \\
\midrule
\multicolumn{7}{l}{\emph{Llama-3.1-8B-Instruct}} \\
QASPER & OPD & 2.6 & 10.7 & 13.1 & 176 & 167 \\
 & APD & 5.5 & 7.1 & \textbf{11.5} & 118 & 172 \\
\cmidrule(lr){1-7}
BioASQ & OPD & 2.5 & 16.0 & 18.4 & 139 & 171 \\
 & APD & 5.7 & 7.6 & \textbf{12.7} & 98 & 176 \\
\midrule
\multicolumn{7}{l}{\emph{Qwen3-8B}} \\
QASPER & OPD & 2.8 & 14.2 & 17.0 & 293 & 170 \\
 & APD & 6.4 & 8.8 & \textbf{14.9} & 249 & 179 \\
\cmidrule(lr){1-7}
BioASQ & OPD & 2.9 & 20.8 & 23.5 & 304 & 175 \\
 & APD & 7.0 & 8.3 & \textbf{15.2} & 249 & 179 \\
\bottomrule
\end{tabular}
\caption{Training cost of OPD and APD on two B200 GPUs (median over 20 optimizer steps after 5 warm-up steps; one document per dataset).
Update includes the separate teacher forward pass of OPD.
Length is the mean rollout length in tokens.
Memory is the peak reserved memory per GPU in GB, including the KV-cache pools that vLLM allocates in advance.}
\label{tab:timing}
\end{table}

\begin{table}[h]
\centering
\small
\setlength{\tabcolsep}{2.5pt}%
\begin{tabular}{lcccccc}
\toprule
& \multicolumn{3}{c}{Llama-3.1-8B-Instruct} & \multicolumn{3}{c}{Qwen3-8B} \\
\cmidrule(lr){2-4}\cmidrule(lr){5-7}
$\tau$ & strict & len. & MRR & strict & len. & MRR \\
\midrule
$10$ & .437 & .472 & .453 & .351 & .466 & .399 \\
$3$ & .449 & .482 & .465 & .353 & .477 & .406 \\
$1$ & .471 & .496 & .483 & .382 & .520 & .440 \\
$0.3$ & \textbf{.512} & \textbf{.550} & \textbf{.530} & .386 & .506 & .436 \\
$0.1$ & .468 & .491 & .478 & \textbf{.394} & \textbf{.538} & \textbf{.454} \\
$0.01$ & .476 & .502 & .489 & .384 & .519 & .440 \\
\bottomrule
\end{tabular}
\caption{BioASQ development scores for selecting $\tau$ (mean over the ten development documents).
The highest value in each column is in \textbf{bold}.}
\label{tab:tau-dev}
\end{table}

\paragraph{Evaluation.} All evaluation is closed-book generation with greedy
decoding (temperature $0$), capped at $128$ new tokens for QASPER and $64$ for BioASQ.
Both reference rows of Table~\ref{tab:main} use the same instruction as the trained models: the \emph{no training} row is the base model on the identical closed-book prompt, and the \emph{base $+$ document} row is that same prompt with the document prepended (Appendix~\ref{sec:appendix-prompts}).

\paragraph{Significance testing.}\label{sec:appendix-significance} Each test document is a separate training run, so we compare two methods with a two-sided paired $t$-test over documents: for each document we take the difference between the two methods' mean scores on its questions, and test whether the mean difference is zero ($n{=}20$, $19$ degrees of freedom).
A difference is reported as significant when $p<.10$, with no correction for multiple comparisons.

\subsection{Temperature Selection}
\label{sec:appendix-tau}

For each model, we select $\tau$ for APD once from $\{0.01, 0.1, 0.3, 1, 3, 10\}$ on ten BioASQ development documents that are disjoint from the twenty test documents, and use the selected value for both datasets.
The training and evaluation sets of these development documents are built by the same pipeline (Appendix~\ref{sec:appendix-datagen}).
We train APD on each development document with each $\tau$ and choose the value with the highest mean strict accuracy.
Table~\ref{tab:tau-dev} lists the development scores, and the selected values, $\tau=0.3$ for Llama-3.1 and $\tau=0.1$ for Qwen3, are the highest on all three metrics.

\subsection{Baseline Settings}
\label{sec:appendix-gate-hparams}

Every baseline uses the method-specific hyperparameters of its original paper, whereas the training setup shared by all methods (optimizer, learning rate, number of epochs, batch size, and maximum length; Appendix~\ref{sec:appendix-realdata}) is ours.
In their original papers, most on-policy baselines first fine-tune the student with SFT and then distill into it.
We instead distill into the student without this step, to measure injection under the largest information asymmetry between student and teacher.
Appendix~\ref{sec:appendix-warmstart} shows that APD remains the best method when every method is first fine-tuned with SFT.
Below we describe the baselines whose settings need comment.

\paragraph{GKD.} We run GKD~\citep{agarwal2024onpolicy} with a student-data fraction of $\lambda=0.5$, which the authors call mixed sampling, and the generalized Jensen--Shannon divergence with $\beta=0.5$.
The original work treats both as task-dependent hyperparameters and tunes them per task.
We did not repeat that search for every setting, but our ablation on Llama-3.1 and BioASQ (Table~\ref{tab:axes}) covers part of it, varying the sampling source and the direction of the KL divergence, though not the Jensen--Shannon divergence.

\paragraph{MiniLLM.} We follow the official implementation of MiniLLM~\citep{gu2024minillm}, including teacher-mixed sampling with $\alpha=0.2$, single-step regularization, and length normalization, with two differences.
We omit the auxiliary language-modeling loss on a pretraining corpus, which our setting does not have.
We perform one policy update per rollout instead of four, as for the other on-policy methods.

\begin{table}[t]
\centering
\small
\setlength{\tabcolsep}{2pt}%
\begin{tabular}{lcccccc}
\toprule
& \multicolumn{6}{c}{SWITCH, decay factor $\lambda$} \\
\cmidrule(lr){2-7}
& $1/5$ & $1/10$ & $1/15$ & $1/50$ & $1/100$ & $1/200$ \\
\midrule
original paper & 78 & 53 & 7 & -- & -- & -- \\
\midrule
Llama, QASPER & 86.0 & 81.0 & 72.8 & 31.2 & 10.7 & 2.2 \\
Llama, BioASQ & 84.2 & 74.4 & 67.5 & 31.3 & 10.9 & 2.4 \\
Qwen, QASPER & 90.6 & 84.4 & 78.7 & 49.6 & 28.5 & 11.5 \\
Qwen, BioASQ & 89.1 & 82.9 & 74.8 & 45.6 & 27.0 & 12.3 \\
\bottomrule
\end{tabular}
\caption{Percentage of rollout tokens generated by the teacher in SWITCH before any update, on one document per dataset.
Our experiments use the original value $\lambda{=}1/10$.
The first row is from Table~4 of \citet{koo2025switch}.}
\label{tab:switch-diag}
\end{table}

\begin{table}[t]
\centering
\small
\setlength{\tabcolsep}{4pt}%
\begin{tabular}{lcccc}
\toprule
& \multicolumn{4}{c}{SKD, acceptance rank $K$} \\
\cmidrule(lr){2-5}
& 25 & 5 & 2 & 1 \\
\midrule
Llama, QASPER & 0.7 & 4.3 & 12.5 & 31.7 \\
Llama, BioASQ & 1.0 & 4.2 & 12.1 & 24.0 \\
Qwen, QASPER & 0.8 & 4.6 & 12.8 & 25.8 \\
Qwen, BioASQ & 0.8 & 3.8 & 11.6 & 22.2 \\
\bottomrule
\end{tabular}
\caption{Percentage of student-sampled tokens that the teacher replaces in SKD before any update, on one document per dataset.
Our experiments use the original value $K{=}25$, and the authors recommend $K$ in $[5,50]$.}
\label{tab:skd-diag}
\end{table}

\paragraph{SWITCH.} SWITCH lets the teacher generate a token whenever the teacher--student divergence exceeds a threshold $\tau_t=\tau_0 e^{-\lambda t}$ that decays with the token position $t$~\citep{koo2025switch}.
We keep the original settings, including $\lambda=1/10$, which the original paper selects as the best-performing value, sampling at temperature $1.0$, and training on rollouts and on the fixed training answers in equal proportion.
Because the threshold decays with position, the proportion of teacher-generated tokens depends on output length, and our rollouts, of up to $512$ tokens, are much longer than the instruction-following outputs of the original work.
At $\lambda=1/10$, the teacher generates $74$--$84\%$ of our rollout tokens before training, more than the $53\%$ that the original paper reports (Table~\ref{tab:switch-diag}).
During training it averages $59$--$76\%$.
Nevertheless, a smaller decay factor does not help.
With $\lambda=1/50$, at which the teacher generates $31$--$50\%$ of rollout tokens, MRR on Llama-3.1 and BioASQ is $.016$ lower ($t=-1.99$), so we keep the original value.

\begin{table*}[p]
\centering
\small
\setlength{\tabcolsep}{5pt}%
\begin{tabular}{l ccc ccc}
\toprule
& \multicolumn{3}{c}{QASPER} & \multicolumn{3}{c}{BioASQ} \\
\cmidrule(lr){2-4}\cmidrule(lr){5-7}
Method & P & R & F1 & strict & lenient & MRR \\
\midrule
\multicolumn{7}{c}{\textbf{Llama-3.1-8B-Instruct}} \\
\midrule
\emph{no training} & .158\std{.046} & .164\std{.044} & .148\std{.040} & .121\std{.074} & .199\std{.097} & .151\std{.082} \\
\emph{base $+$ document} & .756\std{.052} & .860\std{.048} & .764\std{.042} & .733\std{.050} & .769\std{.044} & .748\std{.046} \\
\midrule
\multicolumn{7}{l}{\emph{No teacher resampling during training}} \\
\quad SFT & .475\smash{$^{-}$}\std{.085} & .622\std{.074} & .485\smash{$^{-}$}\std{.070} & .412\smash{$^{-}$}\std{.063} & .488\std{.057} & .446\smash{$^{-}$}\std{.057} \\
\quad PD & .461\smash{$^{-}$}\std{.108} & .520\smash{$^{-}$}\std{.140} & .442\smash{$^{-}$}\std{.108} & .392\smash{$^{-}$}\std{.060} & .407\smash{$^{-}$}\std{.067} & .399\smash{$^{-}$}\std{.063} \\
\quad ToDi & .474\smash{$^{-}$}\std{.146} & .539\smash{$^{-}$}\std{.169} & .460\smash{$^{-}$}\std{.143} & .418\smash{$^{-}$}\std{.078} & .457\smash{$^{-}$}\std{.081} & .435\smash{$^{-}$}\std{.076} \\
\quad GKD & .457\smash{$^{-}$}\std{.112} & .510\smash{$^{-}$}\std{.128} & .440\smash{$^{-}$}\std{.101} & .384\smash{$^{-}$}\std{.059} & .422\smash{$^{-}$}\std{.062} & .400\smash{$^{-}$}\std{.058} \\
\quad OPD & .411\smash{$^{-}$}\std{.153} & .449\smash{$^{-}$}\std{.166} & .390\smash{$^{-}$}\std{.141} & .317\smash{$^{-}$}\std{.062} & .333\smash{$^{-}$}\std{.058} & .324\smash{$^{-}$}\std{.060} \\
\midrule
\multicolumn{7}{l}{\emph{Teacher resampled during training}} \\
\quad \czero & .532\smash{$^{-}$}\std{.112} & .595\std{.130} & .515\smash{$^{-}$}\std{.105} & .451\std{.068} & .482\smash{$^{-}$}\std{.062} & .464\std{.064} \\
\quad MiniLLM & .196\smash{$^{-}$}\std{.076} & .390\smash{$^{-}$}\std{.149} & .221\smash{$^{-}$}\std{.083} & .144\smash{$^{-}$}\std{.053} & .155\smash{$^{-}$}\std{.054} & .149\smash{$^{-}$}\std{.052} \\
\quad SWITCH & .413\smash{$^{-}$}\std{.165} & .436\smash{$^{-}$}\std{.180} & .390\smash{$^{-}$}\std{.158} & .347\smash{$^{-}$}\std{.066} & .376\smash{$^{-}$}\std{.073} & .360\smash{$^{-}$}\std{.068} \\
\quad SKD & .543\smash{$^{-}$}\std{.056} & .610\smash{$^{-}$}\std{.079} & .527\smash{$^{-}$}\std{.053} & .425\smash{$^{-}$}\std{.065} & .447\smash{$^{-}$}\std{.075} & .435\smash{$^{-}$}\std{.069} \\
\quad AdaSwitch & .533\smash{$^{-}$}\std{.067} & .613\std{.076} & .524\smash{$^{-}$}\std{.062} & .413\smash{$^{-}$}\std{.057} & .432\smash{$^{-}$}\std{.060} & .421\smash{$^{-}$}\std{.057} \\
\midrule
\multicolumn{7}{l}{\emph{APD (ours), all values of $\tau$}} \\
\quad $\tau=10$ & .503\smash{$^{-}$}\std{.115} & .536\smash{$^{-}$}\std{.140} & .478\smash{$^{-}$}\std{.112} & .424\smash{$^{-}$}\std{.071} & .472\smash{$^{-}$}\std{.085} & .444\smash{$^{-}$}\std{.073} \\
\quad $\tau=3$ & .566\std{.075} & .599\smash{$^{-}$}\std{.083} & .543\std{.071} & .453\smash{$^{-}$}\std{.071} & .489\smash{$^{-}$}\std{.085} & .469\smash{$^{-}$}\std{.076} \\
\quad $\tau=1$ & .554\std{.121} & .608\std{.143} & .535\std{.118} & .465\std{.071} & .510\std{.087} & .485\std{.075} \\
\quad $\tau=0.3$ & \cellcolor{heat!18}.574\std{.067} & \cellcolor{heat!18}.636\std{.072} & \cellcolor{heat!18}.558\std{.053} & \cellcolor{heat!18}.469\std{.068} & \cellcolor{heat!18}.508\std{.076} & \cellcolor{heat!18}.486\std{.070} \\
\quad $\tau=0.1$ & .574\std{.064} & .625\std{.081} & .551\std{.055} & .463\std{.072} & .493\std{.076} & .476\std{.072} \\
\quad $\tau=0.01$ & .576\std{.054} & .623\std{.070} & .553\std{.043} & .457\std{.061} & .486\std{.067} & .469\std{.062} \\
\midrule[\heavyrulewidth]
\multicolumn{7}{c}{\textbf{Qwen3-8B}} \\
\midrule
\emph{no training} & .140\std{.039} & .250\std{.043} & .157\std{.035} & .115\std{.062} & .209\std{.087} & .151\std{.070} \\
\emph{base $+$ document} & .699\std{.065} & .805\std{.055} & .701\std{.051} & .678\std{.071} & .762\std{.065} & .713\std{.068} \\
\midrule
\multicolumn{7}{l}{\emph{No teacher resampling during training}} \\
\quad SFT & .352\smash{$^{-}$}\std{.060} & .506\smash{$^{-}$}\std{.060} & .373\smash{$^{-}$}\std{.049} & .310\smash{$^{-}$}\std{.075} & .463\smash{$^{-}$}\std{.086} & .373\smash{$^{-}$}\std{.079} \\
\quad PD & .427\smash{$^{-}$}\std{.068} & .489\smash{$^{-}$}\std{.092} & .409\smash{$^{-}$}\std{.060} & .345\smash{$^{-}$}\std{.067} & .457\smash{$^{-}$}\std{.081} & .391\smash{$^{-}$}\std{.071} \\
\quad ToDi & .430\smash{$^{-}$}\std{.062} & .492\smash{$^{-}$}\std{.077} & .415\smash{$^{-}$}\std{.057} & .364\std{.083} & .470\smash{$^{-}$}\std{.098} & .407\smash{$^{-}$}\std{.089} \\
\quad GKD & .378\smash{$^{-}$}\std{.058} & .429\smash{$^{-}$}\std{.062} & .364\smash{$^{-}$}\std{.050} & .317\smash{$^{-}$}\std{.069} & .422\smash{$^{-}$}\std{.069} & .360\smash{$^{-}$}\std{.065} \\
\quad OPD & .309\smash{$^{-}$}\std{.066} & .378\smash{$^{-}$}\std{.075} & .302\smash{$^{-}$}\std{.052} & .244\smash{$^{-}$}\std{.061} & .323\smash{$^{-}$}\std{.076} & .276\smash{$^{-}$}\std{.066} \\
\midrule
\multicolumn{7}{l}{\emph{Teacher resampled during training}} \\
\quad \czero & .447\std{.075} & .509\smash{$^{-}$}\std{.077} & .430\smash{$^{-}$}\std{.060} & .368\std{.078} & .475\smash{$^{-}$}\std{.084} & .413\smash{$^{-}$}\std{.079} \\
\quad MiniLLM & .189\smash{$^{-}$}\std{.072} & .384\smash{$^{-}$}\std{.065} & .218\smash{$^{-}$}\std{.060} & .197\smash{$^{-}$}\std{.063} & .271\smash{$^{-}$}\std{.088} & .228\smash{$^{-}$}\std{.073} \\
\quad SWITCH & .358\smash{$^{-}$}\std{.058} & .378\smash{$^{-}$}\std{.071} & .333\smash{$^{-}$}\std{.048} & .281\smash{$^{-}$}\std{.072} & .379\smash{$^{-}$}\std{.085} & .322\smash{$^{-}$}\std{.073} \\
\quad SKD & .434\smash{$^{-}$}\std{.088} & .507\smash{$^{-}$}\std{.090} & .422\smash{$^{-}$}\std{.072} & .360\std{.071} & .484\std{.104} & .410\std{.081} \\
\quad AdaSwitch & .438\smash{$^{-}$}\std{.075} & .529\std{.083} & .433\smash{$^{-}$}\std{.058} & .359\std{.082} & .476\std{.117} & .408\std{.095} \\
\midrule
\multicolumn{7}{l}{\emph{APD (ours), all values of $\tau$}} \\
\quad $\tau=10$ & .405\smash{$^{-}$}\std{.074} & .458\smash{$^{-}$}\std{.073} & .394\smash{$^{-}$}\std{.064} & .335\smash{$^{-}$}\std{.067} & .451\smash{$^{-}$}\std{.073} & .383\smash{$^{-}$}\std{.067} \\
\quad $\tau=3$ & .418\smash{$^{-}$}\std{.083} & .494\smash{$^{-}$}\std{.077} & .410\smash{$^{-}$}\std{.069} & .379\std{.073} & .488\std{.085} & .423\std{.075} \\
\quad $\tau=1$ & .447\std{.066} & .503\smash{$^{-}$}\std{.085} & .432\std{.061} & .371\std{.064} & .503\std{.091} & .425\std{.074} \\
\quad $\tau=0.3$ & .439\std{.085} & .516\smash{$^{-}$}\std{.090} & .427\smash{$^{-}$}\std{.068} & .369\std{.070} & .491\std{.082} & .419\std{.073} \\
\quad $\tau=0.1$ & \cellcolor{heat!18}.453\std{.073} & \cellcolor{heat!18}.531\std{.084} & \cellcolor{heat!18}.442\std{.058} & \cellcolor{heat!18}.376\std{.071} & \cellcolor{heat!18}.495\std{.079} & \cellcolor{heat!18}.426\std{.073} \\
\quad $\tau=0.01$ & .456\std{.069} & .505\smash{$^{-}$}\std{.087} & .435\std{.063} & .383\std{.065} & .487\std{.074} & .426\std{.066} \\
\bottomrule
\end{tabular}
\caption{Full results (mean\,$\pm$\,standard deviation over the 20 test documents, final checkpoint).
Shaded APD cells use the $\tau$ selected on the development documents (Appendix~\ref{sec:appendix-tau}).
These are the APD values of Table~\ref{tab:main}.
$^{+}$/$^{-}$: significantly above/below the selected APD in the same column (paired $t$-test over documents, $p < .10$).}
\label{tab:full}
\end{table*}

\begin{table}[t]
\centering
\small
\begin{tabular}{lccc}
\toprule
Method & 5 ep & 10 ep & 15 ep \\
\midrule
\multicolumn{4}{l}{\emph{No teacher resampling during training}} \\
\quad SFT & .466 & .446 & .444 \\
\quad PD & .391 & .399 & .402 \\
\quad ToDi & .430 & .435 & .431 \\
\quad GKD & .327 & .400 & .425 \\
\quad OPD & .256 & .324 & .375 \\
\midrule
\multicolumn{4}{l}{\emph{Teacher resampled during training}} \\
\quad \czero & .418 & .464 & .477 \\
\quad MiniLLM & .141 & .149 & .197 \\
\quad SWITCH & .323 & .360 & .375 \\
\quad SKD & .402 & .435 & .461 \\
\quad AdaSwitch & .404 & .421 & .419 \\
\cmidrule(lr){1-4}
\quad APD (ours) & .435 & .486 & .486 \\
\bottomrule
\end{tabular}
\caption{BioASQ MRR of Llama-3.1 trained for 5, 10, and 15 epochs (final checkpoint).
Mean over 20 documents.}
\label{tab:epochs}
\end{table}

\begin{table}[t]
\centering
\small
\setlength{\tabcolsep}{4pt}%
\begin{tabular}{lccc}
\toprule
Loss & fixed corpus & resampled & $\Delta$ ($t$) \\
\midrule
SFT (CE) & .446 & .441 & $-.005$ ($-0.47$) \\
PD (FKL) & .399 & .464 & $\mathbf{+.065}$ ($9.52$) \\
GKD (JSD) & .400 & .399 & $-.001$ ($-0.11$) \\
\bottomrule
\end{tabular}
\caption{Fixed versus resampled teacher corpus (MRR, mean over 20 documents; $t$ from paired tests across documents).
The resampled PD entry is the \czero{} baseline of Table~\ref{tab:main}.}
\label{tab:resample}
\end{table}

\paragraph{SKD.} We use the original hyperparameters of \citet{xu2025skd}, $K=25$, which the authors fix without parameter search, and their sampling settings, temperature $0.5$ with top-$p$ $0.5$ for the student and temperature $0.2$ for the teacher.
In their setting, where a larger teacher distills into a smaller student, the teacher replaces $40.9\%$ of student tokens early in training and about $5\%$ later.
In our setting, the teacher replaces far fewer tokens.
At $K=25$, it replaces at most $1.0\%$ of tokens before any update, and this falls to $0.2$--$0.3\%$ during training.
At $K=5$, the most aggressive value in the range the authors recommend, it replaces at most $4.6\%$.
Only below that range, at $K=2$ and $K=1$, does the rate rise to $12$--$32\%$, which is still below the initial rate of the original paper (Table~\ref{tab:skd-diag}).
This follows from our setting, in which student and teacher are the same model and differ only in whether the document is in the prompt, so $81$--$87\%$ of student-sampled tokens are the teacher's most likely token and $98$--$99.5\%$ are within its top $25$.
Its rollouts are therefore close to those of student sampling with forward KL, the corresponding cell of Table~\ref{tab:axes}, and the two score $.435$ and $.450$ MRR on Llama-3.1 and BioASQ.
Nevertheless, a more aggressive $K$ does not help.
Training with $K=5$ leaves MRR on Llama-3.1 and BioASQ unchanged within noise ($-.008$, $t=-0.78$), so we keep the original value.

\paragraph{AdaSwitch.} We use the original hyperparameters of \citet{peng2025adaswitch}, $k=3$ and a window of $L=10$ tokens, with its sampling settings.
In our setting, 82--90\% of rollouts switch to the teacher, on average around the 16th generated token, and the teacher then generates the rest of the sequence.
The teacher therefore produces 76--94\% of rollout tokens throughout training, the opposite of SKD.

\section{Full Results}
\label{sec:appendix-full}

Table~\ref{tab:full} extends Table~\ref{tab:main} with standard deviations, significance against APD, and APD at every value of $\tau$ we trained.
Table~\ref{tab:gate-mass} gives the gate of APD at each of these values next to its primary score, F1 for QASPER and MRR for BioASQ.

\section{Additional Analyses}
\label{sec:appendix-analyses}

\subsection{Number of Training Epochs}
\label{sec:appendix-epochs}

We train every method on Llama-3.1 and BioASQ for 5, 10, and 15 epochs, keeping all other settings as in \S\ref{sec:exp-data}.
Each length is a separate run with its own cosine schedule, not a snapshot of a longer run.
Table~\ref{tab:epochs} reports MRR over the same twenty documents.
Training on a fixed teacher-generated corpus saturates early: between 5 and 15 epochs PD changes by at most $.011$ and ToDi by at most $.005$, while SFT declines from $.466$ to $.444$.
Resampling the teacher instead keeps adding signal: \czero{} rises from $.418$ to $.477$, so its advantage over PD grows from $+.027$ ($t{=}2.80$) at 5 epochs to $+.065$ ($t{=}9.52$) at 10 and $+.075$ ($t{=}6.45$) at 15, and SKD improves at both steps.
Methods that resample the student also keep improving: OPD at both steps, and GKD mainly between 5 and 10 epochs.
APD gains most between 5 and 10 epochs ($+.051$, $t{=}5.59$) and is unchanged at 15 epochs, so the 10-epoch setting of Table~\ref{tab:main} already reaches its plateau.

\begin{table*}[t]
\centering
\footnotesize
\setlength{\tabcolsep}{3pt}%
\begin{tabular}{l*{16}{c}}
\toprule
& \multicolumn{8}{c}{Llama-3.1-8B-Instruct} & \multicolumn{8}{c}{Qwen3-8B} \\
\cmidrule(lr){2-9}\cmidrule(lr){10-17}
& \multicolumn{4}{c}{QASPER} & \multicolumn{4}{c}{BioASQ} & \multicolumn{4}{c}{QASPER} & \multicolumn{4}{c}{BioASQ} \\
\cmidrule(lr){2-5}\cmidrule(lr){6-9}\cmidrule(lr){10-13}\cmidrule(lr){14-17}
& \multicolumn{3}{c}{gate $\bar g$} & & \multicolumn{3}{c}{gate $\bar g$} & & \multicolumn{3}{c}{gate $\bar g$} & & \multicolumn{3}{c}{gate $\bar g$} & \\
Setting & start & mean & end & F1 & start & mean & end & MRR & start & mean & end & F1 & start & mean & end & MRR \\
\midrule
OPD & .00 & .00 & .00 & .390 & .00 & .00 & .00 & .324 & .00 & .00 & .00 & .302 & .00 & .00 & .00 & .276 \\
\midrule
$\tau{=}10$ & .03 & .03 & .02 & .478 & .03 & .02 & .02 & .444 & .05 & .03 & .02 & .394 & .05 & .03 & .02 & .383 \\
$\tau{=}3$ & .10 & .07 & .05 & .543 & .09 & .06 & .05 & .469 & .14 & .08 & .05 & .410 & .12 & .06 & .05 & .423 \\
$\tau{=}1$ & .25 & .16 & .10 & .535 & .22 & .14 & .09 & .485 & .26 & .15 & .11 & .432 & .23 & .12 & .09 & .425 \\
$\tau{=}0.3$ & .51 & .29 & .21 & .558 & .44 & .26 & .19 & .486 & .42 & .26 & .20 & .427 & .35 & .22 & .16 & .419 \\
$\tau{=}0.1$ & .67 & .43 & .35 & .551 & .56 & .38 & .30 & .476 & .54 & .38 & .32 & .442 & .46 & .32 & .27 & .426 \\
$\tau{=}0.01$ & .82 & .66 & .59 & .553 & .74 & .60 & .54 & .469 & .71 & .62 & .57 & .435 & .61 & .54 & .48 & .426 \\
\bottomrule
\end{tabular}
\caption{Gate temperature in every setting, ordered from OPD by how far the gate opens (mean over 20 documents, final checkpoint). $\bar g$ is the mean gate value, i.e.\ the fraction of teacher mass injected, for the untrained student, averaged over training, and at the end of training.
Every value of $\tau$ is significantly above OPD in every setting (paired $t$-test over documents, $t \geq 2.9$), including $\tau \geq 1$, where the gate averages at most $.16$ over training.}
\label{tab:gate-mass}
\end{table*}

\subsection{Fixed versus Resampled Teacher Corpus}
\label{sec:appendix-resample}

SFT, PD, and GKD train on teacher answers that are generated greedily once before training and reused in every epoch.
We also train each of them on teacher answers sampled at temperature $1.0$ at every step, keeping the loss unchanged.
For PD, this is the \czero{} baseline of Table~\ref{tab:main}.
Table~\ref{tab:resample} shows, for Llama-3.1 on BioASQ, that resampling the teacher helps only with forward KL, raising PD by $.065$ MRR while leaving SFT and GKD unchanged.
This mirrors the testbed of \S\ref{sec:toy}, where the source of the training sequences matters only once the loss is forward KL.

\begin{figure*}[!t]
\input{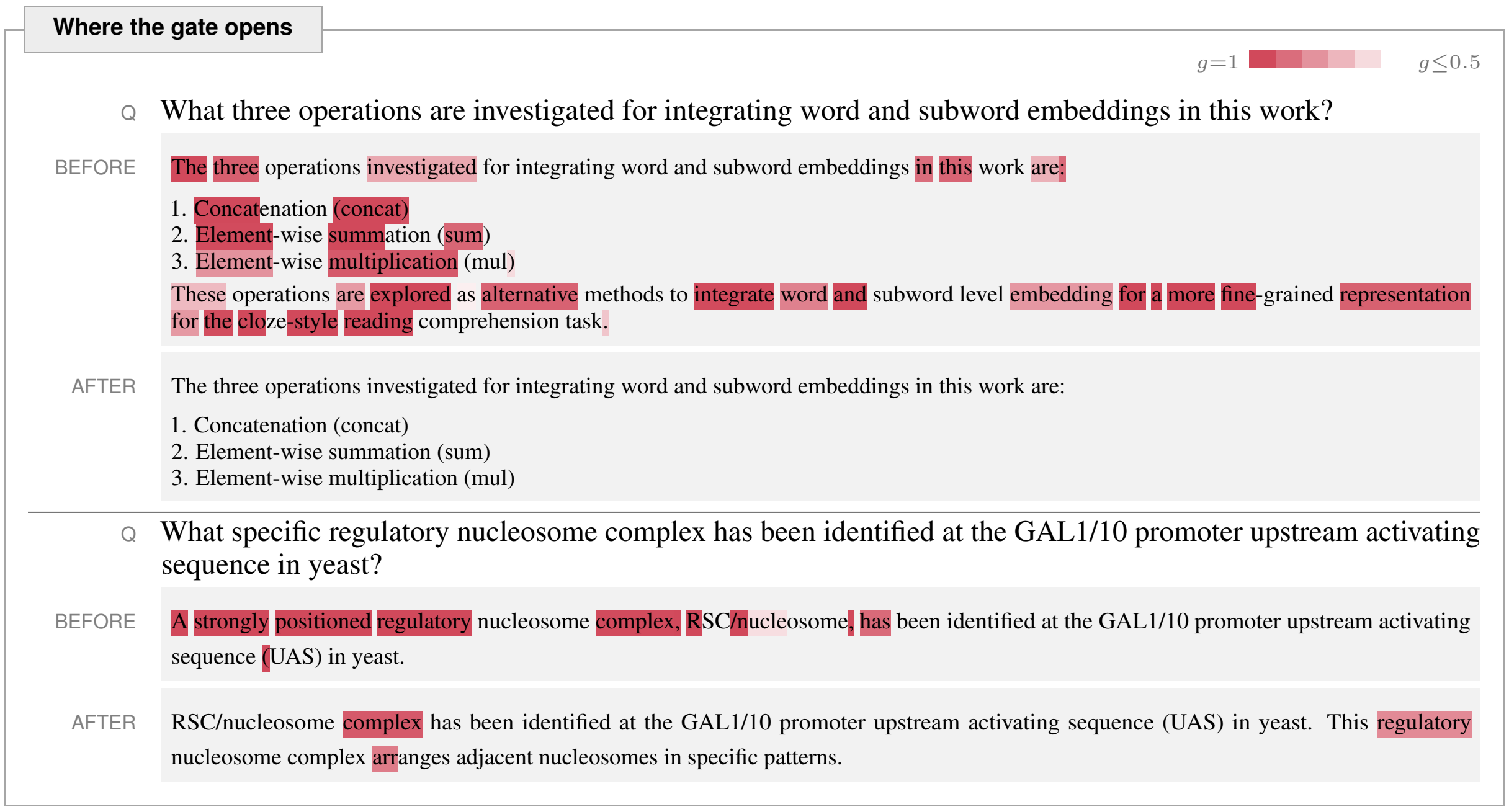}
\caption{Where the gate opens, on two closed-book rollouts (top: QASPER, bottom: BioASQ).
Shading shows the per-token gate of Llama-3.1 before and after 10 epochs of APD ($\tau{=}0.3$), for tokens with $g_t>0.5$: darker means more teacher mass in the sampling mixture and more forward-KL weight in the loss.
Both are \emph{mixture} rollouts, not student-only generations.}
\label{fig:gate-example}
\end{figure*}

\subsection{Gate Examples}
\label{sec:appendix-gate-example}
Figure~\ref{fig:gate-example} shows where the gate opens on two rollouts at the selected $\tau=0.3$.
Before training, it opens on the answer tokens that only the document supplies, such as the three operations of the QASPER paper and the RSC complex in yeast, and on the elaboration that follows them.
It stays closed where the rollout restates the question, such as \textsf{integrating word and subword embeddings} and \textsf{promoter upstream activating sequence}.
After training, the student gives the same answers with the gate closed almost everywhere, and only three words of the BioASQ rollout still receive substantial teacher mass.
The gate thus concentrates on the content that the document must supply and withdraws from it once the student covers it.

\subsection{Degenerate Outputs}
\label{sec:appendix-degenerate}

Mode collapse has been reported for on-policy distillation with reverse KL~\citep{jang2026stable}.
We observe a related failure in our setting, but not only for reverse-KL methods: some trained models give the same answer to many closed-book questions regardless of the question.
On QASPER the repeated answer is ``yes'' or ``no''.
In the most extreme case, $105$ of $106$ answers are ``yes''.
On BioASQ it is an entity name from the article, and all such models were trained on the same document.
Each method yields one trained model per test document, and we count a trained model as degenerate when more than $30\%$ of its answers to its document's questions are the same string after lowercasing and removing trailing punctuation.

Table~\ref{tab:degenerate} counts degenerate models.
They are concentrated in Llama-3.1 on QASPER ($25$ of the $29$ cases) and spread across methods, which suggests that the failure depends more on the model and dataset than on the training method.
APD is not immune: on the same model and dataset, two of its models degenerate at $\tau=1$ and three at $\tau=10$, but none at any $\tau \le 0.3$, including the selected $\tau=0.3$.
The scores in Table~\ref{tab:main} and in all other results include these degenerate models.
When each comparison with APD instead excludes the documents on which either method's model is degenerate, the outcome at $p<.10$ changes in one of $42$ comparisons: the precision of \czero{} for Llama-3.1 on QASPER, where $t$ moves from $-1.73$ to $-1.41$.

\begin{table}[t]
\centering
\small
\setlength{\tabcolsep}{3pt}%
\begin{tabular}{lcccc}
\toprule
& \multicolumn{2}{c}{Llama} & \multicolumn{2}{c}{Qwen} \\
\cmidrule(lr){2-3}\cmidrule(lr){4-5}
Method & QASPER & BioASQ & QASPER & BioASQ \\
\midrule
SFT & 0 & 0 & 0 & 0 \\
PD & 2 & 0 & 0 & 0 \\
ToDi & 3 & 0 & 0 & 0 \\
GKD & 2 & 0 & 0 & 0 \\
OPD & 3 & 0 & 0 & 1 \\
\czero & 2 & 0 & 0 & 0 \\
MiniLLM & 4 & 1 & 0 & 1 \\
SWITCH & 3 & 0 & 0 & 1 \\
SKD & 0 & 0 & 0 & 0 \\
AdaSwitch & 1 & 0 & 0 & 0 \\
\midrule
APD (ours) & 0 & 0 & 0 & 0 \\
\midrule
\multicolumn{5}{l}{\emph{APD, all values of $\tau$}} \\
\quad $\tau=10$ & 3 & 0 & 0 & 0 \\
\quad $\tau=3$ & 0 & 0 & 0 & 0 \\
\quad $\tau=1$ & 2 & 0 & 0 & 0 \\
\quad $\tau=0.3$ & 0 & 0 & 0 & 0 \\
\quad $\tau=0.1$ & 0 & 0 & 0 & 0 \\
\quad $\tau=0.01$ & 0 & 0 & 0 & 0 \\
\bottomrule
\end{tabular}
\caption{Number of degenerate models out of 20 (one per test document), i.e., models for which more than 30\% of the answers are the same string regardless of the question.}
\label{tab:degenerate}
\end{table}

\subsection{Warm-Start (SFT Initialization)}
\label{sec:appendix-warmstart}

GKD initializes on-policy distillation from a supervised fine-tuned checkpoint~\citep{agarwal2024onpolicy}, and SKD reports that on-policy distillation can degrade from a weaker initialization~\citep{xu2025skd}.
To test whether training from the base model disadvantages such methods, we start every method that samples from the student from the 10-epoch SFT checkpoint of the same document and train it for 10 more epochs, keeping the base model as the teacher.
Table~\ref{tab:warmstart} reports the results.

Warm-start helps most where a method is weakest from the base model: OPD gains $+.084$ to $+.131$ and MiniLLM $+.141$ to $+.192$.
APD changes by at most $+.018$ and remains the best method in all four settings.
It is significantly ahead of every other warm-started method in every setting.
APD is also the only method that improves significantly on its SFT initialization in every setting ($+.046$ to $+.082$).
For Llama-3.1, every other method ends at most $.009$ above that initialization, apart from SKD on QASPER ($+.038$).

\begin{table}[t]
\centering
\small
\setlength{\tabcolsep}{3pt}%
\begin{tabular}{lcccc}
\toprule
& \multicolumn{2}{c}{Llama} & \multicolumn{2}{c}{Qwen} \\
\cmidrule(lr){2-3}\cmidrule(lr){4-5}
Method & QASPER & BioASQ & QASPER & BioASQ \\
\midrule
\emph{SFT start} & .485 & .446 & .373 & .373 \\
\emph{APD, from base} & .558 & .486 & .442 & .426 \\
\midrule
GKD & .439\smash{$^{-}$} & .421\smash{$^{-}$} & .433\smash{$^{-}$} & .411\smash{$^{-}$} \\
OPD & .494\smash{$^{-}$} & .409\smash{$^{-}$} & .416\smash{$^{-}$} & .407\smash{$^{-}$} \\
MiniLLM & .385\smash{$^{-}$} & .340\smash{$^{-}$} & .401\smash{$^{-}$} & .370\smash{$^{-}$} \\
SWITCH & .485\smash{$^{-}$} & .413\smash{$^{-}$} & .391\smash{$^{-}$} & .398\smash{$^{-}$} \\
SKD & .523\smash{$^{-}$} & .423\smash{$^{-}$} & .440\smash{$^{-}$} & .424\smash{$^{-}$} \\
AdaSwitch & .462\smash{$^{-}$} & .429\smash{$^{-}$} & .439\smash{$^{-}$} & .420\smash{$^{-}$} \\
\midrule
APD (ours) & .559 & .492 & .455 & .443 \\
\bottomrule
\end{tabular}
\caption{Warm-start: each method trained for 10 epochs from the 10-epoch SFT checkpoint of the same document, with the base model as teacher (QASPER F1, BioASQ MRR; mean over 20 documents). $^{+}$/$^{-}$: significantly above/below warm-started APD in the same column (paired $t$-test over documents, $p < .10$).}
\label{tab:warmstart}
\end{table}

\begin{table*}[p]
\centering
\small
\setlength{\tabcolsep}{6pt}%
\begin{tabular}{l ccccccc}
\toprule
Method & IFEval & MATH & HumanEval & MMLU-Pro & BBH & GPQA & Avg. \\
\midrule
\multicolumn{8}{c}{\textbf{Llama-3.1-8B-Instruct}, trained on QASPER} \\
\midrule
\emph{no training} & .827 & .204 & .689 & .458 & .668 & .255 & .517 \\
\midrule
\quad SFT & .745 & .133 & .580 & .393 & .555 & .261 & .444 \\
\quad PD & .560 & .129 & .559 & .360 & .539 & .254 & .400 \\
\quad ToDi & .610 & .141 & .577 & .389 & .540 & .255 & .419 \\
\quad GKD & .601 & .138 & .573 & .365 & .536 & .250 & .411 \\
\quad OPD & .595 & .136 & .575 & .346 & .511 & .250 & .402 \\
\quad \czero & .606 & .138 & .560 & .375 & .551 & .244 & .412 \\
\quad MiniLLM & .564 & .135 & .585 & .332 & .486 & .248 & .392 \\
\quad SWITCH & .605 & .135 & .571 & .368 & .536 & .243 & .410 \\
\quad SKD & .585 & .130 & .563 & .341 & .536 & .250 & .401 \\
\quad AdaSwitch & .591 & .135 & .574 & .355 & .541 & .249 & .408 \\
\cmidrule(lr){1-8}
\quad APD (ours) & .605 & .140 & .566 & .376 & .553 & .247 & .415 \\
\midrule[\heavyrulewidth]
\multicolumn{8}{c}{\textbf{Llama-3.1-8B-Instruct}, trained on BioASQ} \\
\midrule
\emph{no training} & .827 & .204 & .689 & .458 & .668 & .255 & .517 \\
\midrule
\quad SFT & .752 & .128 & .585 & .380 & .544 & .253 & .440 \\
\quad PD & .580 & .123 & .566 & .339 & .527 & .255 & .398 \\
\quad ToDi & .628 & .137 & .591 & .377 & .545 & .258 & .423 \\
\quad GKD & .642 & .138 & .602 & .348 & .539 & .245 & .419 \\
\quad OPD & .626 & .136 & .589 & .327 & .531 & .245 & .409 \\
\quad \czero & .621 & .140 & .583 & .364 & .552 & .248 & .418 \\
\quad MiniLLM & .553 & .127 & .595 & .288 & .484 & .240 & .381 \\
\quad SWITCH & .621 & .131 & .596 & .359 & .529 & .246 & .414 \\
\quad SKD & .603 & .125 & .591 & .326 & .535 & .247 & .404 \\
\quad AdaSwitch & .602 & .128 & .596 & .342 & .542 & .255 & .411 \\
\cmidrule(lr){1-8}
\quad APD (ours) & .643 & .138 & .593 & .362 & .552 & .253 & .424 \\
\midrule[\heavyrulewidth]
\multicolumn{8}{c}{\textbf{Qwen3-8B}, trained on QASPER} \\
\midrule
\emph{no training} & .871 & .526 & .841 & .623 & .846 & .441 & .691 \\
\midrule
\quad SFT & .829 & .522 & .822 & .593 & .784 & .400 & .658 \\
\quad PD & .845 & .515 & .841 & .592 & .810 & .408 & .669 \\
\quad ToDi & .852 & .525 & .837 & .615 & .807 & .410 & .674 \\
\quad GKD & .852 & .514 & .848 & .607 & .809 & .400 & .672 \\
\quad OPD & .853 & .513 & .850 & .613 & .805 & .387 & .670 \\
\quad \czero & .841 & .502 & .834 & .593 & .805 & .399 & .662 \\
\quad MiniLLM & .850 & .513 & .845 & .606 & .809 & .392 & .669 \\
\quad SWITCH & .857 & .518 & .831 & .601 & .810 & .401 & .670 \\
\quad SKD & .844 & .508 & .843 & .591 & .808 & .396 & .665 \\
\quad AdaSwitch & .844 & .507 & .835 & .593 & .809 & .395 & .664 \\
\cmidrule(lr){1-8}
\quad APD (ours) & .844 & .506 & .834 & .598 & .809 & .394 & .664 \\
\midrule[\heavyrulewidth]
\multicolumn{8}{c}{\textbf{Qwen3-8B}, trained on BioASQ} \\
\midrule
\emph{no training} & .871 & .526 & .841 & .623 & .846 & .441 & .691 \\
\midrule
\quad SFT & .842 & .541 & .822 & .604 & .796 & .407 & .668 \\
\quad PD & .837 & .521 & .846 & .581 & .815 & .414 & .669 \\
\quad ToDi & .852 & .533 & .848 & .614 & .818 & .412 & .679 \\
\quad GKD & .848 & .522 & .851 & .592 & .824 & .407 & .674 \\
\quad OPD & .851 & .520 & .848 & .600 & .815 & .396 & .671 \\
\quad \czero & .837 & .515 & .840 & .579 & .817 & .408 & .666 \\
\quad MiniLLM & .848 & .521 & .852 & .544 & .809 & .392 & .661 \\
\quad SWITCH & .855 & .522 & .841 & .597 & .820 & .402 & .673 \\
\quad SKD & .843 & .516 & .852 & .561 & .819 & .399 & .665 \\
\quad AdaSwitch & .838 & .516 & .840 & .575 & .814 & .402 & .664 \\
\cmidrule(lr){1-8}
\quad APD (ours) & .844 & .513 & .838 & .573 & .819 & .407 & .666 \\
\bottomrule
\end{tabular}
\caption{General capability after injection, per benchmark and averaged over the six (mean over the 20 models trained on each dataset; IFEval instruction-level strict accuracy, HumanEval pass@1, exact-match accuracy otherwise).}
\label{tab:capability}
\end{table*}

\section{Additional Evaluations}
\label{sec:appendix-evals}

\subsection{General Capability}
\label{sec:appendix-capability}

Injecting a document may degrade abilities unrelated to it.
We evaluate every trained model of Table~\ref{tab:main} on six generation benchmarks with the LM Evaluation Harness~\citep{gao2024lmeval}: IFEval~\citep{zhou2023ifeval}, the hardest level of MATH~\citep{hendrycks2021math}, HumanEval~\citep{chen2021codex}, a stratified subset of $1{,}500$ MMLU-Pro questions~\citep{wang2024mmlupro}, BBH~\citep{suzgun2023bbh}, and GPQA~\citep{rein2024gpqa}, the last three with chain-of-thought prompting, five-shot for MMLU-Pro and zero-shot for the other two.
Table~\ref{tab:capability} reports each benchmark and the mean of the six.

Every method loses capability, Llama-3.1 far more than Qwen3: from $.517$ to between $.381$ and $.444$, against $.691$ to between $.658$ and $.679$.
For Llama-3.1, SFT retains the most ($.444$ and $.440$).
Among the distillation methods, APD ($.415$ and $.424$) is the highest on BioASQ and within $.004$ of the highest, ToDi, on QASPER.
For Qwen3, all methods lie within $.021$ of one another, and APD is at most $.013$ below the best of them.
APD thus loses general capability to about the same degree as the other distillation methods.

\subsection{Human-Written Benchmark Questions}
\label{sec:appendix-native}

Each benchmark supplies only a few human-written questions per document: $62$ answerable QASPER questions over $19$ of the $20$ test documents, as the remaining document has only unanswerable ones, and one BioASQ question per document.
Table~\ref{tab:native} reports them, scored per document and averaged as in Table~\ref{tab:main}.
With so few questions, most differences between methods are not significant.
Table~\ref{tab:examples} shows one such question per dataset.
Many QASPER questions presuppose the paper as context, such as \emph{how are rare words defined?}, which is ambiguous without knowing the paper, so we add the paper title to these questions (Appendix~\ref{sec:appendix-prompts}).

On QASPER, APD improves on the untrained model by $+.213$ F1 for Llama-3.1 and $+.174$ for Qwen3, and no method is significantly above it.
On BioASQ, the untrained models already reach $.392$ and $.500$ MRR, close to the open-book $.487$ and $.588$, so these questions largely test knowledge that the base models already hold and separate the methods little.
Apart from ToDi on Qwen3 BioASQ, no method is significantly above APD in any setting.

\begin{table}[t]
\centering
\small
\setlength{\tabcolsep}{2pt}%
\begin{tabular}{lcccc}
\toprule
& \multicolumn{2}{c}{Llama} & \multicolumn{2}{c}{Qwen} \\
\cmidrule(lr){2-3}\cmidrule(lr){4-5}
Method & QASPER & BioASQ & QASPER & BioASQ \\
\midrule
\emph{no training} & .240 & .392 & .204 & .500 \\
\emph{base $+$ document} & .494 & .487 & .469 & .588 \\
\midrule
SFT & .471 & .500 & .382 & .625 \\
PD & .384 & .500 & .375 & .633 \\
ToDi & .376 & .417\smash{$^{-}$} & .375 & .700\smash{$^{+}$} \\
GKD & .388 & .467 & .377 & .617 \\
OPD & .356 & .300\smash{$^{-}$} & .282\smash{$^{-}$} & .625 \\
\czero & .373 & .500 & .400 & .525 \\
MiniLLM & .305\smash{$^{-}$} & .200\smash{$^{-}$} & .281 & .533 \\
SWITCH & .320\smash{$^{-}$} & .500 & .328 & .475 \\
SKD & .465 & .550 & .379 & .600 \\
AdaSwitch & .449 & .550 & .389 & .592 \\
\midrule
APD (ours) & .453 & .550 & .378 & .592 \\
\bottomrule
\end{tabular}
\caption{The benchmarks' human-written questions (QASPER F1 over 62 answerable questions from 19 documents; BioASQ MRR over one question per document, 20 documents). $^{+}$/$^{-}$: significantly above/below APD in the same column (paired $t$-test over documents, $p < .10$).}
\label{tab:native}
\end{table}

\section{Prompts}
\label{sec:appendix-prompts}

Every prompt used in this work is reproduced below.
Italic braces mark a slot filled at run time.
The content is verbatim, and only the line wrapping is cosmetic.
Each prompt is given as a system turn (where one is given) plus a single user turn, and thinking mode is disabled for Qwen3.

\subsection{Training-Data Generation}

We first give the single-passage question-generation prompt.
The cross-passage and full-document scopes use the same prompt with the passage description and rules 2--4 replaced as noted below.

\begin{promptbox}[title={Questions: system prompt (single-passage scope)}]
You curate a question dataset about the given article. Each question's subject is the article's content --- its facts, claims, findings or events, entities, values, relationships, sequences, and outcomes. You are shown one passage from a longer article; treat it as a factual source. Questions must stand alone.\\[3pt]
Rules:\\
1. Self-contained questions --- STRICTLY no meta-reference to your prompt. A reader who has read the whole article but does NOT see your prompt must be able to answer. Forbidden phrasings include but are not limited to: "the excerpt", "the passage", "the article", "the source", "the text", "the section", "the piece", "the paragraph", "this excerpt", "this passage", "this article", "in the passage", "in the excerpt", "according to the excerpt", "according to the passage", "the author writes / says / states / argues / describes". Rephrase by naming the subject directly. Bad: "What does the article say about Dole?" Good: "What did Dole charge the New York Times with?" --- questions read like trivia about the topic, not like reading-comprehension prompts.\\
2. Scope each question so the answer is fully determined by facts in the passage you are shown.\\
3. Vary the angle: factual lookup, what/which/who/where/when/how, comparison, cause, condition, mechanism, outcome.\\
4. Evidence: 1-2 sentence verbatim quote from the passage that supports the answer.\\
5. Write all questions in English.\\
6. FORM DIVERSITY: vary the question FORM widely --- beyond what/which/how, also use comparison, ordering/sequence, cause-effect, fill-in-the-blank, true/false, 'which is NOT', counting, multi-step reasoning, and multiple-choice (3-4 options).\\
7. ANSWER-FORMAT CUES: for SOME questions whose answer is a short exact fact, embed an explicit answer instruction inside the question itself --- e.g. 'answer in N words or fewer', 'give only the short answer, no explanation', 'reply yes or no', 'answer as a comma-separated list'. Leave others open (why/how) for a fuller answer. Aim for a real spread of expected answer lengths across the set.\\
Output format: JSON array with items of shape \{"question": "...", "evidence": "..."\}.
\end{promptbox}

\noindent At \emph{cross}-passage scope the opening becomes ``You are shown
several sections from a longer article''.

\begin{promptbox}[title={Cross-passage scope: replaces rules 2--4}]
2. Each question must require synthesizing content from TWO OR MORE sections; nothing answerable from a single section alone.\\
3. Favor comparison, sequencing, cause-and-effect, or contrast across sections.\\
4. Evidence: 1-2 sentence verbatim quote from the sections that supports the answer.
\end{promptbox}

\noindent At \emph{full}-document scope the subject becomes ``the article as a
whole --- its overall aim, central findings or events, main claims, big-picture relationships'', and the model is shown the full article.

\begin{promptbox}[title={Full-document scope: replaces rules 2--4}]
2. Favor big-picture questions requiring whole-article understanding: the overall aim, central findings, the main comparison, the conclusion.\\
3. Prefer questions that cannot be answered from a single paragraph.\\
4. Evidence: 1-2 sentence verbatim quote that supports the answer.
\end{promptbox}

\begin{promptbox}[title={Questions: user turn (single-passage scope)}]
Generate \pvar{n} distinct, self-contained questions whose answers are facts in the article passage below. Phrase each question as if it were trivia about the topic --- never refer to "the passage" / "the excerpt" / "the article" / "the text" or any other prompt-context label.\\[3pt]
-{}-{}-\\
\pvar{passage}\\
-{}-{}-\\[3pt]
Generate \pvar{n} questions in English.
\end{promptbox}

\begin{promptbox}[title={Answers: user turn (no system prompt)}]
<context>\\
\pvar{document}\\
</context>\\[3pt]
\pvar{question}
\end{promptbox}

\subsection{Evaluation-Set Generation}

\begin{promptbox}[title={System prompt (QASPER)}]
You write question-answer pairs answerable ONLY from the given article. Each question must name a specific entity/finding/value from the article (self-contained), not a bare relative reference. Answer each question in the SAME concise style and length as the EXAMPLE pairs. Output ONLY a JSON array of \{"question": str, "answers": [str]\}, no prose.
\end{promptbox}

\begin{promptbox}[title={System prompt (BioASQ)}]
You write factoid question-answer pairs answerable ONLY from the given article. Each question must name a specific entity/finding/value from the article (self-contained), not a bare relative reference, and its answer must be a SINGLE named entity or short canonical term (a gene, protein, disease, method, or number) exactly as it appears in the article --- NOT a descriptive phrase, clause, or sentence. Answer each question in the SAME concise style as the EXAMPLE pairs. Output ONLY a JSON array of \{"question": str, "answers": [str]\}, no prose.
\end{promptbox}

\noindent The example block repeats for up to five pairs.

\begin{promptbox}[title={User turn (both datasets)}]
ARTICLE:\\
\pvar{article}\\[3pt]
EXAMPLE (question, answer) pairs --- match this style and length:\\
- Q: \pvar{human-written question}\\
~~A: \pvar{human-written answer}\\[3pt]
Generate 100 diverse, non-redundant (question, answer) pairs answerable only from this article, each naming a specific entity. JSON array of \{"question": str, "answers": [str]\}.
\end{promptbox}

\subsection{Training-Time Inputs}

\begin{promptbox}[title={Teacher user turn}]
<context>\\
\pvar{document}\\
</context>\\[3pt]
\pvar{question}
\end{promptbox}

\begin{promptbox}[title={Student user turn}]
\pvar{question}
\end{promptbox}

\subsection{Evaluation}

\begin{promptbox}[title={QASPER, closed-book}]
Answer the question as concisely as you can, using a single phrase or sentence if possible. If the question is a yes/no question, answer "yes" or "no". Do not provide any explanation.\\[3pt]
Question: \pvar{question}\\[3pt]
Answer:
\end{promptbox}

\noindent This is the form used on the generated questions.
On the human-written questions the same prompt carries the paper title inside the question, whose first letter is lower-cased to fit the clause:

\begin{promptbox}[title={QASPER, closed-book: human-written questions (question line only)}]
Question: According to the paper "\pvar{title}", \pvar{question}
\end{promptbox}

\begin{promptbox}[title={BioASQ, closed-book (both question sets)}]
Answer the question with up to five short answers --- a single word or short phrase each --- ranked from most likely to least likely and separated by semicolons. Give only the answers, with no explanation.\\[3pt]
Question: \pvar{question}\\[3pt]
Answer:
\end{promptbox}

\noindent In both datasets the open-book reference row of
Table~\ref{tab:main} uses the same prompt as above with \texttt{Article: \pvar{document}} prepended.
No other wording changes.

\end{document}